\documentclass[letterpaper]{article} 
\usepackage{aaai2026}  
\usepackage{times}  
\usepackage{helvet}  
\usepackage{courier}  
\usepackage[hyphens]{url}  
\usepackage{graphicx} 
\usepackage{natbib}  
\usepackage{caption} 
\usepackage{algorithm}
\usepackage{algorithmic}

\usepackage[table]{xcolor}
\usepackage{subcaption} 
\usepackage{multirow}
\usepackage{booktabs}

\usepackage{mdframed}

\usepackage{amsmath}
\usepackage{amssymb}
\usepackage{amsthm}

\newtheorem{definition}{Definition}

\newtheorem{theorem}{Theorem}
\newtheorem{assumption}{Assumption}
\usepackage{amsfonts}

\usepackage{newfloat}
\usepackage{listings}
\DeclareCaptionStyle{ruled}{labelfont=normalfont,labelsep=colon,strut=off} 
\floatstyle{ruled}
\newfloat{listing}{tb}{lst}{}
\floatname{listing}{Listing}
\nocopyright
\title{LetheViT: Selective Machine Unlearning for Vision Transformers via Attention-Guided Contrastive Learning}
\author{
    Yujia Tong\textsuperscript{\rm 1},
    Tian Zhang\textsuperscript{\rm 1},
    Jingling Yuan\textsuperscript{\rm 1} \thanks{Corresponding author.},
    Yuze Wang\textsuperscript{\rm 1},
    Chuang Hu\textsuperscript{\rm 2}
}
\affiliations{
    \textsuperscript{\rm 1}School of Computer Science and Artificial Intelligence, Wuhan University of Technology\\

    \textsuperscript{\rm 2}State Key Laboratory of Internet of Things for Smart City, University of Macau

    \{tyjjjj, zhangttt, yjl,  hbtmwyz\}@whut.edu.cn, chuanghu@um.edu.mo
}

\usepackage{bibentry}

\begin{document}

\maketitle

\begin{abstract}
Vision Transformers (ViTs) have revolutionized computer vision tasks with their exceptional performance. However, the introduction of privacy regulations such as GDPR and CCPA has brought new challenges to them. These laws grant users the right to withdraw their data, necessitating not only the deletion of data but also the complete removal of its influence from trained models. Machine unlearning emerges as a critical solution, with exact unlearning being computationally prohibitive and approximate methods offering a more practical approach. This work addresses the particularly challenging scenario of random data forgetting in ViTs, where the model must forget specific samples while retaining others, even within the same class. We first reveal the core characteristics of ViTs through selective masking experiments: when high-attention areas are masked, the model retains its recognition capability but significantly weakens its memorization ability. Based on the above insights, we propose LetheViT, a contrastive unlearning method tailored for ViTs. LetheViT uses masked image inputs to generate positive logits and original image inputs to generate negative logits, guiding the model to forget specific details while retaining the general cl category outlines. Experimental results demonstrate that LetheViT achieves state-of-the-art performance, effectively balancing privacy compliance with model efficacy.
\end{abstract}

\section{Introduction}
Privacy regulations such as the General Data Protection Regulation (GDPR) \cite{hoofnagle2019european} and the California Consumer Privacy Act (CCPA) \cite{nguyen2022empirical} have introduced new challenges for Vision Transformers (ViTs) \cite{dosovitskiy2020image}. These laws grant users the right to withdraw their personal data, a withdrawal that requires not only erasing the data from all storage systems but also eliminating every trace of its influence on the model’s training. \textit{Machine Unlearning} (MU) \cite{tong2025robustmachineunlearningquantized,  fan2024simplicity,liu2024model} using a reverse-learning process to erase the impact of specific data points on the model and thereby safeguard user privacy emerges as a promising solution. The most direct and effective method of MU is \textit{Exact Unlearning} \cite{bourtoule2021machine}, which retrains a new model from scratch using the remaining training set. However, this approach requires a substantial amount of computational resources. To address this challenge, \textit{Approximate Unlearning} has been proposed, which can eliminate the impact of specific data without retraining from scratch \cite{chien2022certified}.

Based on the degree of forgetting, MU can be divided into two types: 1) \textit{Class-wise Forgetting} \cite{liu2024model}, which removes all data of a specific class from the model. For example, if a law bans recognizing a particular political symbol on a social media platform using a ViT for content moderation, the model must forget all images labeled with that symbol to comply. In this case, most existing approximate unlearning methods can achieve performance comparable to exact unlearning. 2) \textit{Random Data Forgetting} \cite{tong2025robustmachineunlearningquantized}, which involves forgetting randomly selected samples from one or more classes. For example, on the same social media platform, users may request the removal of their specific images (e.g., pictures of their pet ``cat") from the training data due to privacy concerns.  The model must forget these images while still recognizing other users' content in the same class. This is more common in real-world applications. \textit{We focus on Random Data Forgetting.}

However, compared with class-wise forgetting, random data forgetting significantly increases the complexity, resulting in a substantial performance gap between existing approximate unlearning methods and exact unlearning. The core challenge lies in the need to precisely ``erase" individual samples within the same class while retaining other highly similar samples. For example, in the ``cat" class, if two nearly indistinguishable images are present—one in the forget set and the other in the retain set—directly performing a forgetting operation on the model will weaken the forgetting effect. More critically, existing methods \cite{golatkar2020eternal, liu2024model} generally overlook the unique characteristics of the self-attention mechanism in ViTs. When these methods are directly applied to ViTs, the forgetting effect is further diminished.

To address this challenge, we first explore the recognition and memorization capabilities of ViT models through systematic experiments on selective patch masking. Specifically, we mask the highest-attention patches identified via self-attention scores and evaluate the model's test accuracy (TA) and membership inference attack (MIA) success rate (The lower the MIA success rate, the more difficult it is for the model to distinguish whether a data sample was used in training). Our key observation reveals a critical phenomenon: masking 5\% of top-attended patches with zero pixels preserves recognition capability (TA increases by 0.01\%) while significantly degrading memorization (MIA drops by 14.33\%). This indicates that ViTs retain class-level abstraction when critical details are obscured, yet lose sample-specific memory traces.

Based on the above insights, we propose LetheViT, a novel contrastive unlearning method specifically designed for ViT models. Specifically, samples from the forget set are first passed through the original model to obtain the logits of the negative set. Then, after masking the key information (First, the most important tokens are identified, and then the corresponding image pixels of these tokens are set to 0) in these samples, they are forwarded through the original model again to obtain the logits of the positive set. Meanwhile, the samples in the forgetting set are also passed through the unlearned model to obtain the logits of the anchor.
During the unlearning training process, the goal is to adjust the logits of the anchor so that they are closer to the logits of the positive set in terms of feature representation, while being farther away from the logits of the negative set. This type of contrastive unlearning enables the ViT model to forget the specific details of certain samples within a class while retaining a general outline of the category. In this way, it achieves selective forgetting of particular samples.
We summarize our contributions below: 

\begin{itemize}
\item  We analyze the challenges of ViT models in random forgetting scenarios and explore the Recognition and Memorization capabilities of ViT models.
\item  We propose LetheViT, a machine unlearning method specifically designed for ViT models, which achieves the forgetting of specific samples while retaining the model's performance on the retained dataset.
\item  We conduct extensive experiments to verify our method. The experimental results show that our method achieves state-of-the-art performance compared with existing methods.
\end{itemize}

\section{Preliminaries}
In this section, we revisit the basic concepts of Vision Transformer, machine unlearning, and contrastive learning.

\noindent\textbf{Revisiting Vision Transformer.}
In a typical vision transformer, the input is processed as a sequence of vectors. The process begins by dividing the input image into a fixed number of uniformly sized patches. Each patch is then linearly transformed into a vector. These vectors, referred to as tokens, are input into the vision transformer as \( X \). The token vectors \( X \) pass through several transformer blocks, each consisting of a multi-head self-attention (MSA) module followed by a multi-layer perceptron (MLP) module. For each attention head, the attention weights are calculated using \( Q_i = X W^{Q}_i \), \( K_i = X W^{K}_i \), and \( V_i = X W^{V}_i \), with the attention mechanism defined as:
\begin{equation}
  \text{Att}_i(Q_i, K_i, V_i) = \text{softmax}\left(Q_i K^T_i/\sqrt{d}\right) V_i ,
\label{eq1}
\end{equation}
where \( d \) represents the hidden dimension of each head. The outputs from all heads are combined through concatenation to form the MSA output:
\begin{equation}
  \text{MSA}(X) = \text{concat}(\text{Att}_1, \text{Att}_2, \ldots, \text{Att}_i)W,
\label{eq2}
\end{equation}
where \( i \) denotes the number of heads. The MSA output is then fed into the MLP.

\noindent\textbf{Revisiting Machine Unlearning.}
Let the complete training dataset be \( D = \{(x_i, y_i)\}_{i=1}^N \), consisting of \( N \) samples, where \( x_i\) denotes the \( i \)-th sample and \( y_i \in \{1, 2, \dots, n\} \) is its associated class label. The forget set \( D_f \subseteq D \) represents a subset of \( D \) that needs to be removed from the trained model, while its complement, the retain set \( D_r \), contains the data to be preserved, satisfying \( D_f \cap D_r = \varnothing \) and \( D_f \cup D_r = D \). Machine unlearning (MU) in image classification can be categorized based on the composition of \( D_f \): class-wise forgetting and random data forgetting. In class-wise forgetting, \( D_f \) comprises solely samples from a single class, with the objective of eliminating the influence of that entire class on the model. In random data forgetting, \( D_f \) includes randomly selected samples from one or multiple classes, aiming to remove their impact on the model. Prior to unlearning, the original model is denoted as \( f_{\theta_o} \). In MU, the retrained model \( f_{\theta_r} \), trained from scratch on \( D_r \), is considered the ``gold standard''~\cite{nguyen2022survey,tong2025robustmachineunlearningquantized}. However, retraining incurs significant computational overhead. To address this, approximate unlearning aims to produce an unlearned model \( f_{\theta_u} \) by removing the influence of \( D_f \) from \( f_{\theta_u} \), thereby approximating \( f_{\theta_r} \) with reduced computational cost.

\noindent\textbf{Revisiting Contrastive Learning.}
Contrastive learning aims to learn effective representations by comparing pairs of samples in a dataset \( D \). The dataset \( D \) contains samples \( x_i\), each associated with a class label \( y_i \in \{1, 2, \dots, k\} \). The objective is to train a model \( f \) to map samples into a feature space where positive pairs, typically formed by augmenting a sample \( x_i \) to create \( x_i^+ \), are positioned closely together, while negative pairs, derived from samples \( x_i^- \) of different classes or unrelated data, are placed far apart. The model \( f \) optimizes a loss function that maximizes the similarity between representations \( \mathcal{Z} = f(x_i) \) and \( \mathcal{Z}_p = f(x_i^+) \) for positive pairs, while minimizing similarity with representations \( \mathcal{Z}_n = f(x_i^-) \) for negative pairs. A similarity metric, such as cosine similarity, and a temperature parameter are used to control the distribution's softness, enhancing the model's ability to distinguish between similar and dissimilar samples in the feature space.

\section{Exploring the Memory and Recognition Abilities of ViTs}

To systematically investigate the memory and recognition capabilities of ViTs, we conducted a series of experiments where we selectively masked the most attended image patches—identified through the highest attention scores—and evaluated the model's performance in terms of test accuracy (TA) and the success rate of membership inference attacks (MIA). The memory and recognition capabilities of ViTs are defined as follows:

\begin{definition}[Recognition Capability of ViTs]
In the task of image classification, the recognition capability of a Vision Transformer (i.e., its ability to accurately identify and classify visual patterns in unseen data) is typically reflected by the model's Top-1 classification accuracy on the test set (test accuracy, TA).
\end{definition}

\begin{definition}[Memorization Capacity of ViTs]
The memory capacity of a Vision Transformer reflects its degree of memorization of training data. In image classification tasks, this ability can be quantitatively evaluated through the success rate of Membership Inference Attacks (MIA) — that is, the probability that the model successfully identifies the true training status (whether or not it belongs to the training data) of data samples on the forgetting set.
\end{definition}

Specifically, we conduct experiments on the CIFAR-100 dataset using the DeiT-T model, with a forgetting scenario of randomly forgetting 10\% of the data. We first train a retrain model using the retained data. Subsequently, we apply different masking ratios to the images in the test set and the forgetting set. We then measure the model's Top-1 classification accuracy on the test set (i.e., test accuracy, TA) and the success rate of membership inference attacks (MIA) on the forgetting set. As shown in Table \ref{tab1}, we employ two masking methods: setting pixels to zero and applying Gaussian noise. When the masking ratio is 5\%, we find that setting pixels to zero actually increases the TA by 0.01\% compared to using the original images. This indicates that even after masking the pixels corresponding to the highest-attention patches, the model can still recognize the class of the image, and its recognition ability remains intact. Moreover, the success rate of MIA drops from 24.49\% to 10.16\%, suggesting that the model finds it more difficult to determine whether the masked image was used for training, indicating a significant reduction in the model's memory of the image.

\begin{figure}[t]
\centering
\includegraphics[width=0.472\textwidth]{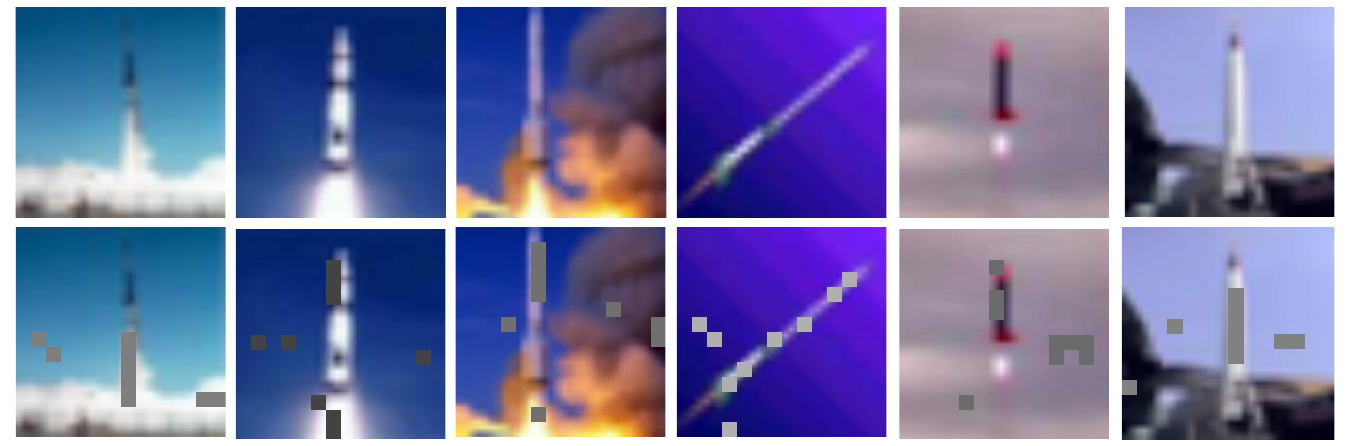}
\caption{Visualization of the original image and the masked image (with 5\% masking). The class is “rocket.”}
\label{fig1}
\end{figure}

We present the masked images and the original images in Figure \ref{fig1}. Taking the ``rocket” class as an example, the masked patches mainly cover the detailed parts of the rocket, while the main outline of the rocket is still preserved. As a result, the model can still correctly identify the class of the masked image. However, due to the introduction of a small amount of noise, the model finds it difficult to determine whether the masked image was used for training. Based on the above insights, we will introduce our method LetheViT in the next section.

\begin{table}[t]
\centering
\small
\setlength{\tabcolsep}{1.5mm}{
\begin{tabular}{c|cc|cc}
\toprule 
\multirow{2}{*}{Ratio} & \multicolumn{2}{c|}{Zero Noise} & \multicolumn{2}{c}{Gaussian Noise} \\ 
\cmidrule(lr){2-3} \cmidrule(lr){4-5}
 & TA & MIA & TA & MIA \\ 
\midrule
$0\%$   & 81.24 & 24.49 & 81.24 & 24.49 \\ 
$5\%$   & $81.25^{\textcolor[RGB]{34,149,34}{\uparrow0.01}}$ & $10.16^{\textcolor[RGB]{221,0,1}{\downarrow14.33}}$ & $83.59^{\textcolor[RGB]{34,149,34}{\uparrow2.35}}$ & $14.06^{\textcolor[RGB]{221,0,1}{\downarrow10.43}}$ \\ 
$10\%$  & $79.69^{\textcolor[RGB]{221,0,1}{\downarrow1.55}}$ & $18.75^{\textcolor[RGB]{221,0,1}{\downarrow5.74}}$ & $81.25^{\textcolor[RGB]{34,149,34}{\uparrow0.01}}$ & $19.53^{\textcolor[RGB]{221,0,1}{\downarrow4.96}}$ \\ 
$20\%$  & $68.75^{\textcolor[RGB]{221,0,1}{\downarrow12.49}}$ & $38.24^{\textcolor[RGB]{34,149,34}{\uparrow13.75}}$ & $69.53^{\textcolor[RGB]{221,0,1}{\downarrow11.71}}$ & $37.50^{\textcolor[RGB]{34,149,34}{\uparrow13.01}}$ \\ 
$30\%$  & $53.91^{\textcolor[RGB]{221,0,1}{\downarrow27.33}}$ & $52.34^{\textcolor[RGB]{34,149,34}{\uparrow27.85}}$ & $57.81^{\textcolor[RGB]{221,0,1}{\downarrow23.43}}$ & $56.25^{\textcolor[RGB]{34,149,34}{\uparrow31.76}}$ \\ 
\bottomrule
\end{tabular}
}
\caption{TA and MIA  under different masking ratios.}
\label{tab1}
\end{table}

\section{Methodology}

We propose a novel contrastive unlearning approach tailored for Vision Transformers (ViTs), enabling selective forgetting of designated samples while preserving performance on retained samples. Our method leverages the attention mechanism to identify and mask critical image regions, guiding the model to forget targeted information through a contrastive learning framework.

\begin{figure*}[t]
\centering
\includegraphics[width=1\textwidth]{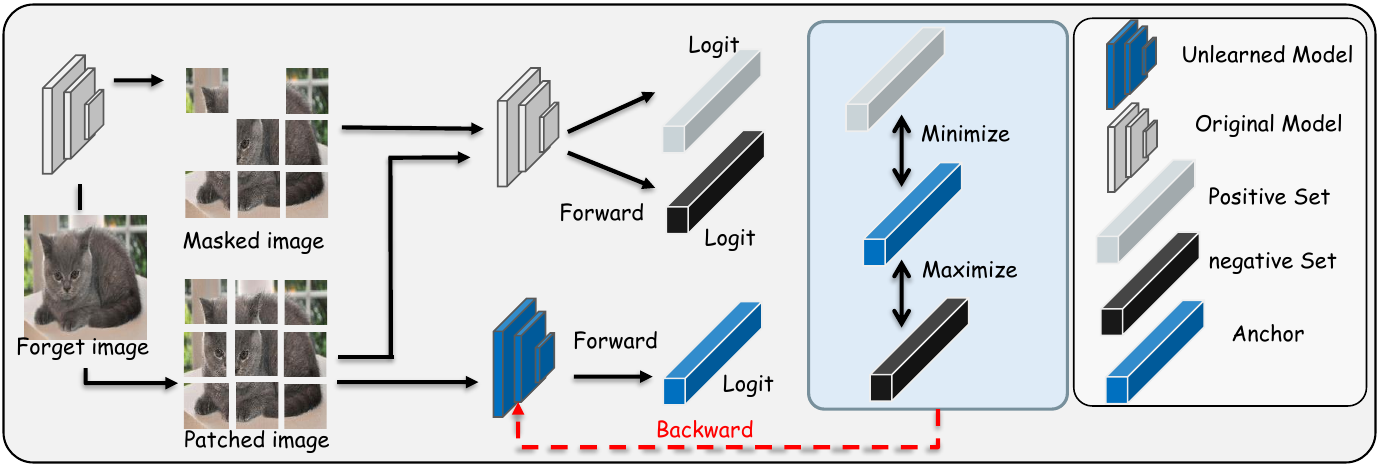}
\caption{The overview of LetheViT.}
\label{fig2}
\end{figure*}

\subsection{Attention-Guided Masking}
For a given input image \( x \), we extract the attention maps from the last attention layer of the original pre-trained model. These attention maps, denoted as \( A \in \mathbb{R}^{B \times H \times (N+1) \times (N+1)} \), where \( B \) is the batch size, \( H \) is the number of attention heads, and \( N \) is the number of patches (with an additional class token), represent the pairwise attention weights between tokens.

To identify the most informative patches, we compute the attention weight from the class token to each patch:
\begin{equation}
a_i = \frac{1}{H} \sum_{h=1}^{H} A_{h,0,i+1} ,
\label{eq3}
\end{equation}
for each patch \( i \) (\( 1 \leq i \leq N \)), where \( A_{h,0,i+1} \) is the attention weight from the class token (position 0) to patch token \( i \) (position \( i+1 \)) in head \( h \). We select the top \( k \) patches with the highest attention scores \( a_i \), where \( k = \lfloor \rho \cdot N \rfloor \), and \( \rho \) is the masking ratio. The masked image \( x_m \) is created by setting the pixel values of these selected patches to zero.

\subsection{Contrastive Unlearning Loss}

In the previous section, we explored the recognition and memorization capabilities of ViT: after masking the patch with the highest attention score in an image, ViT can still identify the class of the image, but has difficulty determining whether the image was used for training. Based on these insights, we can implement selective forgetting through contrastive learning. Specifically, we denote \( f_{\theta_u} \) as the current model being unlearned and \( f_{\theta_o} \) as the original pre-trained model. For an input image \( x \), we first compute the anchor and the positive and negative sets as follows:
\begin{equation}
 \mathcal{Z} = f_{\theta_u}(x), \mathcal{Z}_p = f_{\theta_o}(x_m),   \mathcal{Z}_n = f_{\theta_o}(x) ,
 \label{eq4}
\end{equation}
where \( \mathcal{Z} \) is the representation of the original image on the current model, \( \mathcal{Z}_p \) is the representation of the masked image on the original model, and \( \mathcal{Z}_n \) is the representation of the original image on the original model. The contrastive loss is defined as:

\begin{equation}
\mathcal{L}_{\text{CL}} = -\log \frac{\exp(\text{sim}(\mathcal{Z}, \mathcal{Z}_p) / \tau)}{\exp(\text{sim}(\mathcal{Z}, \mathcal{Z}_p) / \tau) + \exp(\text{sim}(\mathcal{Z}, \mathcal{Z}_n) / \tau)},
\label{eq5}
\end{equation}
where \( \text{sim}(\cdot, \cdot) \) is the cosine similarity, and \( \tau \) is a temperature parameter. This loss encourages the current model's representation \( \mathcal{Z} \) to be closer to \(\mathcal{Z}_p \) and farther from \( \mathcal{Z}_n \). By optimizing this loss function, the model can remember class-level features while forgetting detailed features, thereby achieving selective forgetting of specific samples.

\begin{algorithm}[t]
\caption{The Overall Pipeline of Unlearning}
\textbf{Input:} Pre-trained ViT model \( f_{\theta_o} \) with parameters \(\theta_o\), Forget set \( D_f = \{(x_f, y_f)\} \), Retain set \( D_r = \{(x_r, y_r)\} \),Forget Set Training Epochs \( E_f \), Retain Set Training Epochs \( E_r \),  Learning rate \( \eta \).\\
\textbf{Output:} Unlearned model \( f_{\theta_u} \) .
\begin{algorithmic}[1] 
\STATE Initialize \( \theta_u \leftarrow \theta_0 \)
\FOR{$t \in [0, ..., E_f-1]$}
                \STATE Mask image \( (x_f, y_f) \)  in forget set following Eq.(\ref{eq3})
                \STATE Compute \( \mathcal{Z}, \mathcal{Z}_p, \mathcal{Z}_n \) following Eq.(\ref{eq4})  
                \STATE Compute \( \mathcal{L}_{\text{\textit{CL}}}  \)  following Eq.(\ref{eq5}) 
            \STATE Update \( \theta_u \leftarrow \theta_u - \eta \nabla \mathcal{L}_{\text{\textit{CL}}} \) 
    \ENDFOR 
    \FOR{$t \in [0, ..., E_r-1]$}
                \STATE Compute Cross-Entropy loss \( \mathcal{L}_{\text{\textit{CE}}} = \text{CE}(f_{\theta_u}(x_r), y_r) \)
            \STATE Update \( \theta_u \leftarrow \theta_u - \eta \nabla \mathcal{L}_{\text{\textit{CE}}} \)
    \ENDFOR 
\STATE \textbf{Return} Unlearned model \( f_{\theta_u} \)  with parameters \( \theta_u \)
\end{algorithmic}
\label{alg1}
\end{algorithm}

\subsection{The Overall Pipeline of Unlearning}

We present the overall pipeline of unlearning in Algorithm \ref{alg1}. The unlearning process of LetheViT involves training the model by processing the forget and retain sets. Specifically, during the forget set training phase, for each batch in the forget set, we compute the contrastive loss \( \mathcal{L}_{\text{\textit{CL}}} \) and update the model parameters \( \theta_u \) to achieve selective forgetting of specific samples.  During the retain set training phase, for each batch in the retain set, we compute the classification loss \( \mathcal{L}_{\text{\textit{CE}}} \) and update the model parameters \( \theta_u \) to maintain the model's performance on the retain set. Training iterations on the forget and retain sets is predetermined before training.

\subsection{Theoretical Analysis}
We analyze the efficacy of the LetheViT from the perspective of mutual information.

\begin{mdframed}[backgroundcolor=gray!15,hidealllines=true,innerleftmargin=2pt,innerrightmargin=2pt]
\begin{assumption}
For attention-guided masking, we have:
\begin{equation}
I(\mathcal{Z}_p; \mathcal{S} \mid \mathcal{C}) \approx 0 
\quad \text{and} \quad 
I(\mathcal{Z}_n; \mathcal{S} \mid \mathcal{C}) > 0,
\label{eq6}
\end{equation}
where $I(\cdot;\cdot\mid\cdot)$ denotes conditional mutual information, $\mathcal{C}$ represents class-level information, and $\mathcal{S}$ denotes sample-specific details targeted for forgetting.
\end{assumption}
\end{mdframed}
Let \( \mathcal{Z}_p = f_{\theta_0}(x_m) \) denote the representation of the masked input \( x_m \), which primarily encodes class information \( \mathcal{C} \). We can assume that \( \mathcal{Z}_p \) is conditionally independent of \( \mathcal{S} \) given \( \mathcal{C} \), yielding \( I(\mathcal{Z}_p; \mathcal{S} \mid \mathcal{C}) \approx 0 \). In contrast, the representation of the unmasked input, \( \mathcal{Z}_n = f_{\theta_0}(x) \), encodes both \( \mathcal{C} \) and \( \mathcal{S} \), resulting in \( I(\mathcal{Z}_n; \mathcal{S} \mid \mathcal{C}) > 0 \). The empirical results (Table \ref{tab1}, Figure \ref{fig1}) show that attention-guided masking can effectively remove high-attention regions containing sample-specific details \( \mathcal{S} \) while preserving the class-level contours encoded in \( \mathcal{C} \). The model retains robust recognition performance but exhibits negligible signs of memorizing sample-specific information. This demonstrates that the assumption holds in common settings.

\begin{mdframed}[backgroundcolor=gray!15,hidealllines=true,innerleftmargin=2pt,innerrightmargin=2pt]
\begin{theorem}
Minimizing the loss $\mathcal{L}_{\text{CL}}$ is equivalent to:
\begin{equation}
\max I(\mathcal{Z};\mathcal{Z}_p)
\quad \text{and} \quad
\min I(\mathcal{Z};\mathcal{Z}_n),
\label{eq7}
\end{equation}
\end{theorem}
\end{mdframed}

\begin{proof}
The contrastive loss can be expressed by Eq.(5), which corresponds to the InfoNCE loss with a single negative sample ($\mathcal{K}=1$). From information theory \cite{oord2018representation}, we have:
\begin{equation}
I(\mathcal{Z};\mathcal{Z}_p)\geq \log(\mathcal{K})-\mathcal{L}_{\text{\textit{CL}}} = -\mathcal{L}_{\text{\textit{CL}}},
\label{eq8}
\end{equation}
since $\mathcal{K}=1$. Thus, minimizing $\mathcal{L}_{\text{\textit{CL}}}$ maximizes the lower bound on the mutual information $I(\mathcal{Z};\mathcal{Z}_p)$. Concurrently, the term $\exp\!\bigl(\operatorname{sim}(\mathcal{Z},\mathcal{Z}_n)/\tau\bigr)$ in the denominator drives a reduction in $I(\mathcal{Z};\mathcal{Z}_n)$, as it penalizes similarity between $\mathcal{Z}$ and $\mathcal{Z}_n$.
\end{proof}

\begin{mdframed}[backgroundcolor=gray!15,hidealllines=true,innerleftmargin=2pt,innerrightmargin=2pt]
\begin{theorem}
Under Assumption 1, optimizing the contrastive loss $\mathcal{L}_{\text{\textit{CL}}}$ ensures that the unlearned model satisfies:
\begin{equation}
I(\mathcal{Z};\mathcal{S}\mid\mathcal{C})\to 0,
\label{eq9}
\end{equation}
effectively forgetting sample-specific details $\mathcal{S}$ while preserving class-level information $\mathcal{C}$.
\end{theorem}
\end{mdframed}

\begin{proof}
Combining the results from Assumption 1 and Theorem ~1, we establish the following:
\begin{itemize}
  \item \textbf{Maximizing $I(\mathcal{Z};\mathcal{Z}_p)$:} 
    Since $\mathcal{Z}_p$ depends solely on class information $\mathcal{C}$, the data-processing inequality implies $I(\mathcal{Z};\mathcal{Z}_p)\leq I(\mathcal{Z};\mathcal{C})$. Thus, maximizing $I(\mathcal{Z};\mathcal{Z}_p)$ ensures that $I(\mathcal{Z};\mathcal{C})$ is preserved, maintaining class-level knowledge.
  \item \textbf{Minimizing $I(\mathcal{Z};\mathcal{Z}_n)$:} 
    As $\mathcal{Z}_n$ encodes both $\mathcal{C}$ and $\mathcal{S}$, and assuming $\mathcal{S}$ is conditionally independent of $\mathcal{C}$ in the random forgetting scenario, we have:
    \begin{equation}
    I(\mathcal{Z};\mathcal{Z}_n)\geq I(\mathcal{Z};\mathcal{S}\mid\mathcal{C}).
    \label{eq10}
    \end{equation}
    since $\mathcal{Z}_n$ contains information about $\mathcal{S}$. Reducing $I(\mathcal{Z};\mathcal{Z}_n)$ consequently diminishes $I(\mathcal{Z};\mathcal{S}\mid\mathcal{C})$.
\end{itemize}
Thus, optimizing $\mathcal{L}_{\text{\textit{CL}}}$ directly reduces $I(\mathcal{Z};\mathcal{S}\mid\mathcal{C})$, enabling the model to forget sample-specific details $\mathcal{S}$ while retaining class information $\mathcal{C}$.
\end{proof}

\begin{table*}[t]
  \centering
  \small
  \setlength{\tabcolsep}{0.8mm}{
    \begin{tabular}{c|ccccc|ccccc}
      \toprule
      \multirow{2}[4]{*}{Method} & \multicolumn{5}{c|}{\textbf{ViT-T}} & \multicolumn{5}{c}{\textbf{ViT-S}} \\
      \cmidrule(lr){2-6} \cmidrule(lr){7-11}
       & FA & RA & TA & MIA & AG$\color{blue}\downarrow$ & FA & RA & TA & MIA & AG$\color{blue}\downarrow$ \\
      \midrule
      Retrain &78.89&95.77&79.58&35.78    &\cellcolor{gray!20}0  &86.52&99.57&86.32&24.15&\cellcolor{gray!20}0 \\
      FT &80.43\textcolor{blue}{(1.54)} & 87.57\textcolor{blue}{(8.26)} &80.68\textcolor{blue}{(1.10)} &37.78\textcolor{blue}{(2.00)} &3.23 \cellcolor{gray!20} &84.18\textcolor{blue}{(2.24)}&99.00\textcolor{blue}{(0.57)}&82.42\textcolor{blue}{(3.90)}&30.13\textcolor{blue}{(5.58)}&\underline{3.07} \cellcolor{gray!20} \\
      GA &76.10\textcolor{blue}{(2.79)}&76.98\textcolor{blue}{(18.79)}&74.87\textcolor{blue}{(4.71)}&47.72\textcolor{blue}{(11.94)}&9.56 \cellcolor{gray!20} &96.76\textcolor{blue}{(10.24)}&96.93\textcolor{blue}{(2.64)}&87.62\textcolor{blue}{(1.30)}&12.57\textcolor{blue}{(11.58)}&6.44 \cellcolor{gray!20} \\
      IU &71.35\textcolor{blue}{(7.54)}&73.49\textcolor{blue}{(22.28)}&71.07\textcolor{blue}{(8.51)}&46.62\textcolor{blue}{(10.84)}&12.29 \cellcolor{gray!20} &96.27\textcolor{blue}{(9.75)}&96.71\textcolor{blue}{(2.86)}&87.34\textcolor{blue}{(1.02)}&13.29\textcolor{blue}{(10.86)}&6.12 \cellcolor{gray!20} \\
      RL &79.50\textcolor{blue}{(0.61)} &87.20\textcolor{blue}{(8.57)} & 80.16\textcolor{blue}{(0.58)} &36.93\textcolor{blue}{(1.15)} &2.73 \cellcolor{gray!20} &90.99\textcolor{blue}{(4.47)}&98.75\textcolor{blue}{(0.82)}&86.80\textcolor{blue}{(0.48)}&33.72\textcolor{blue}{(9.57)}&3.84         \cellcolor{gray!20} \\
      $\ell_1$-sparse & 80.54\textcolor{blue}{(0.56)} & 89.00\textcolor{blue}{(6.77)} &80.72\textcolor{blue}{(1.14)} & 36.37\textcolor{blue}{(0.59)} &\underline{2.27} \cellcolor{gray!20} &84.10\textcolor{blue}{(2.42)}&98.98\textcolor{blue}{(0.59)}&82.22\textcolor{blue}{(4.10)}&30.79\textcolor{blue}{(6.64)}&3.43
      \cellcolor{gray!20} \\
      SalUn  &79.45\textcolor{blue}{(0.56)}&89.38\textcolor{blue}{(6.39)}&80.04\textcolor{blue}{(0.46)}&38.63\textcolor{blue}{(2.85)} &2.56 \cellcolor{gray!20} &88.01\textcolor{blue}{(1.49)}&98.72\textcolor{blue}{(0.85)}&86.28\textcolor{blue}{(0.04)}&35.07\textcolor{blue}{(10.92)}&3.33 \cellcolor{gray!20} \\
      LetheViT & 80.09\textcolor{blue}{(1.20)} & 91.55\textcolor{blue}{(4.22)} &80.26\textcolor{blue}{(0.68)} &36.81\textcolor{blue}{(1.03)} & \textbf{1.78} \cellcolor{gray!20} &91.14\textcolor{blue}{(4.62)} &98.79\textcolor{blue}{(0.78)} &85.92\textcolor{blue}{(0.40)} &22.04\textcolor{blue}{(2.11)}&\textbf{2.00} \cellcolor{gray!20} \\
      \midrule
      & \multicolumn{5}{c|}{\textbf{ViT-B}} & \multicolumn{5}{c}{\textbf{DeiT-T}} \\
      \midrule
      Retrain &87.62&99.98&87.92&23.89&\cellcolor{gray!20}0 &76.53&91.65&76.92&40.06&\cellcolor{gray!20}0  \\
      FT  &83.05 \textcolor{blue}{(4.57)} &99.74 \textcolor{blue}{(0.24)} & 81.02\textcolor{blue}{(6.90)} &30.51 \textcolor{blue}{(6.62)} &4.58 \cellcolor{gray!20} &86.55\textcolor{blue}{(10.02)}&96.48\textcolor{blue}{(4.83)}&75.76\textcolor{blue}{(1.16)}&30.93\textcolor{blue}{(9.13)}&6.29 \cellcolor{gray!20}\\
      GA &99.99\textcolor{blue}{(12.37)}&99.96\textcolor{blue}{(0.02)}&88.62\textcolor{blue}{(0.70)}&1.87\textcolor{blue}{(22.02)}&8.78 \cellcolor{gray!20} &93.40\textcolor{blue}{(16.87)}&93.27\textcolor{blue}{(1.62)}&77.46\textcolor{blue}{(0.54)}&22.90\textcolor{blue}{(17.16)}&9.05 \cellcolor{gray!20}\\
      IU &99.99\textcolor{blue}{(12.37)}&99.95\textcolor{blue}{(0.03)}&88.22\textcolor{blue}{(0.30)}&2.35\textcolor{blue}{(21.54)}&8.56 \cellcolor{gray!20}  &90.64\textcolor{blue}{(14.11)}&91.58\textcolor{blue}{(0.07)}&75.70\textcolor{blue}{(1.22)}&22.44\textcolor{blue}{(17.62)} &8.26\cellcolor{gray!20} \\
      RL & 94.50\textcolor{blue}{(6.88)} &99.97\textcolor{blue}{(0.01)} & 86.74\textcolor{blue}{(1.18)} &51.01\textcolor{blue}{(27.12)} &8.80
      \cellcolor{gray!20} &85.26\textcolor{blue}{(8.73)} &95.79\textcolor{blue}{(4.14)} &76.24\textcolor{blue}{( 0.68)} &36.18\textcolor{blue}{(3.88)} &4.36 \cellcolor{gray!20}\\
      $\ell_1$-sparse&83.07\textcolor{blue}{(4.55)} & 99.72\textcolor{blue}{(0.26)} &80.82 \textcolor{blue}{(7.10)} &29.48\textcolor{blue}{(5.59)} &\underline{4.38} \cellcolor{gray!20} &86.55\textcolor{blue}{(10.02)}&96.50\textcolor{blue}{(4.85)}&75.90\textcolor{blue}{(1.02)}&30.79\textcolor{blue}{(9.27)} &6.29\cellcolor{gray!20}\\
      SalUn &96.94\textcolor{blue}{(9.32)} &99.95\textcolor{blue}{(0.03)} &87.12\textcolor{blue}{(0.80)} &38.20\textcolor{blue}{(14.31)} &6.12 \cellcolor{gray!20} &84.05\textcolor{blue}{(7.52)}&95.18\textcolor{blue}{(3.53)}&75.66\textcolor{blue}{(1.26)}&35.79\textcolor{blue}{(4.27)}&\underline{4.15}\cellcolor{gray!20}\\
      LetheViT &86.03\textcolor{blue}{(1.59)} &99.19\textcolor{blue}{(0.79)} &82.54\textcolor{blue}{(5.35)} &25.06\textcolor{blue}{(1.17)} & \textbf{2.23} \cellcolor{gray!20} &80.90\textcolor{blue}{(4.37)} &94.09\textcolor{blue}{(2.44)} &75.04\textcolor{blue}{(1.88)} &37.60\textcolor{blue}{(2.46)} &\textbf{2.79}\cellcolor{gray!20} \\
      \midrule
      & \multicolumn{5}{c|}{\textbf{DeiT-S}} & \multicolumn{5}{c}{\textbf{DeiT-B}} \\
      \midrule
      Retrain &85.00 &99.34 &85.58 &25.83 &0 \cellcolor{gray!20} &90.54 &99.95 &90.60 &19.95 &0 \cellcolor{gray!20} \\
      FT &89.96\textcolor{blue}{(4.96)} &98.71\textcolor{blue}{(0.63)} &86.12\textcolor{blue}{(0.54)} &22.17\textcolor{blue}{(3.66)} &\underline{2.45}\cellcolor{gray!20} &93.02\textcolor{blue}{(2.48)} &99.51\textcolor{blue}{(0.44)} &90.70\textcolor{blue}{(0.10)} &17.01\textcolor{blue}{(2.94)} &1.49\cellcolor{gray!20}\\
      GA &93.42\textcolor{blue}{(8.42)} &93.69\textcolor{blue}{(5.65)} &87.02\textcolor{blue}{(1.44)} &22.20\textcolor{blue}{(3.63)} & 4.79\cellcolor{gray!20} &95.06\textcolor{blue}{(4.52)} &95.50\textcolor{blue}{(4.45)} &91.62\textcolor{blue}{(1.02)} &16.32\textcolor{blue}{(3.63)} &3.41\cellcolor{gray!20}\\
      IU &92.48\textcolor{blue}{(7.48)} &93.21\textcolor{blue}{(6.13)} &86.42\textcolor{blue}{(0.84)} &23.76\textcolor{blue}{(2.07)} & 4.13\cellcolor{gray!20} &94.82\textcolor{blue}{(4.28)} &95.40\textcolor{blue}{(4.55)} &91.38\textcolor{blue}{(0.78)} &16.92\textcolor{blue}{(3.03)} & 3.16\cellcolor{gray!20}\\
      RL &85.16\textcolor{blue}{(0.16)} &98.41\textcolor{blue}{(0.93)} &85.70\textcolor{blue}{(0.12)} &40.73\textcolor{blue}{(14.90)} & 4.03\cellcolor{gray!20} &90.22\textcolor{blue}{(0.32)} &99.04\textcolor{blue}{(0.91)} &91.20\textcolor{blue}{(0.60)} &25.41\textcolor{blue}{(5.46)} &1.82\cellcolor{gray!20}\\
      $\ell_1$-sparse &82.17\textcolor{blue}{(2.83)} &99.14\textcolor{blue}{(0.20)} &80.78\textcolor{blue}{(4.80)} &31.97\textcolor{blue}{(6.14)} &3.49\cellcolor{gray!20} &88.15\textcolor{blue}{(2.39)} &99.74\textcolor{blue}{(0.21)} &87.70\textcolor{blue}{(2.90)} &24.59\textcolor{blue}{(4.64)} &2.54\cellcolor{gray!20}\\
      SalUn &85.77\textcolor{blue}{(0.77)} &97.82\textcolor{blue}{(1.52)} &85.32\textcolor{blue}{(0.26)} &33.32\textcolor{blue}{(7.49)} &2.51\cellcolor{gray!20} &90.31\textcolor{blue}{(0.23)} &98.86\textcolor{blue}{(1.09)} &91.00\textcolor{blue}{(0.40)} &23.25\textcolor{blue}{(3.30)} &\underline{1.26}\cellcolor{gray!20}\\
      LetheViT & 87.87\textcolor{blue}{(2.87)} &98.31\textcolor{blue}{(1.03)} &85.34\textcolor{blue}{(0.24)} &25.56\textcolor{blue}{(0.27)} & \textbf{1.10}\cellcolor{gray!20} &90.70\textcolor{blue}{(0.16)} &99.40\textcolor{blue}{(0.55)} &89.38\textcolor{blue}{(1.22)} &20.28\textcolor{blue}{(0.33)} & \textbf{0.57}\cellcolor{gray!20}\\
      \midrule
      & \multicolumn{5}{c|}{\textbf{Swin-T}} & \multicolumn{5}{c}{\textbf{Swin-S}} \\
      \midrule
      Retrain &84.89&99.13&85.56&26.36&\cellcolor{gray!20}0 &88.22&99.92&88.72&21.08&\cellcolor{gray!20}0 \\
      FT &78.92\textcolor{blue}{(5.97)} &96.99\textcolor{blue}{(2.14)} & 78.68\textcolor{blue}{(6.88)} &35.15\textcolor{blue}{(8.79)} &5.95\cellcolor{gray!20} &81.74\textcolor{blue}{(6.48)} &98.50 \textcolor{blue}{(1.42)} & 80.36\textcolor{blue}{(8.36)} &30.87\textcolor{blue}{(9.79)} &6.51\cellcolor{gray!20}\\
      GA&96.38\textcolor{blue}{(11.49)}&96.41\textcolor{blue}{(2.72)}&87.18\textcolor{blue}{(1.62)}&14.35\textcolor{blue}{(12.01)}&6.96\cellcolor{gray!20}  &98.94\textcolor{blue}{(10.72)}&99.02\textcolor{blue}{(0.90)}&89.42\textcolor{blue}{(0.70)}&5.94\textcolor{blue}{(15.14)}&6.87 \cellcolor{gray!20}\\
      IU &90.13\textcolor{blue}{(5.24)}&91.42\textcolor{blue}{(7.71)}&82.80\textcolor{blue}{(2.76)}&23.08\textcolor{blue}{(3.28)}&4.75\cellcolor{gray!20} &98.80\textcolor{blue}{(10.58)}&98.91\textcolor{blue}{(1.01)}&88.80\textcolor{blue}{(0.08)}&6.63\textcolor{blue}{(14.45)}&6.53\cellcolor{gray!20}\\
      RL  &88.64\textcolor{blue}{(3.75)} &98.40 \textcolor{blue}{(0.73)} & 86.74\textcolor{blue}{(1.18)} &38.65 \textcolor{blue}{(12.29)} &4.49\cellcolor{gray!20}  &83.12\textcolor{blue}{(5.10)}&99.52\textcolor{blue}{(0.40)}&86.00\textcolor{blue}{(2.72)}&54.35\textcolor{blue}{(33.27)}&10.37\cellcolor{gray!20}\\
      $\ell_1$-sparse &80.85 \textcolor{blue}{(4.04)} &97.90 \textcolor{blue}{(1.23)} &79.82\textcolor{blue}{(5.74)} & 33.05\textcolor{blue}{(6.69)} &4.43\cellcolor{gray!20} & 81.55\textcolor{blue}{(6.67)} & 98.37\textcolor{blue}{(1.55)} &80.82 \textcolor{blue}{(7.90)} &31.14\textcolor{blue}{(10.06)} &6.55\cellcolor{gray!20}\\
      SalUn  & 87.76\textcolor{blue}{(2.87)} &98.11\textcolor{blue}{(1.02)} & 85.64\textcolor{blue}{(0.08)} &38.29\textcolor{blue}{(11.93)} &\underline{3.98}\cellcolor{gray!20} &88.06\textcolor{blue}{(0.16)}&99.37\textcolor{blue}{(0.55)}&87.10\textcolor{blue}{(1.62)}&35.00\textcolor{blue}{(13.92)}&\underline{4.06}\cellcolor{gray!20}\\
      LetheViT & 88.53\textcolor{blue}{(3.64)} &98.16 \textcolor{blue}{(0.97)} &84.66 \textcolor{blue}{(0.90)} &25.12 \textcolor{blue}{(1.24)} & \textbf{1.69}\cellcolor{gray!20} &91.50\textcolor{blue}{(3.28)} &99.27\textcolor{blue}{(0.65)} &86.14\textcolor{blue}{(2.58)} &19.74\textcolor{blue}{(1.34)} &\textbf{1.96}\cellcolor{gray!20}\\
      \bottomrule
    \end{tabular}
  }
  \caption{Performance of various MU methods on Tiny-Imagenet. \textbf{Bold} indicates the best performance and \underline{underline} indicates the runner-up. A performance gap against Retrain is provided in \textcolor{blue}{(•)}. The proportion of forgotten data samples is 10\%.}
  \label{tab2}
\end{table*}

\section{Experiments}

\subsection{Experimental Setup}
\noindent\textbf{Datasets and Networks.} The datasets used in the experiments are CIFAR-10 \cite{krizhevsky2009learning}, CIFAR-100 \cite{krizhevsky2009learning}, SVHN \cite{netzer2011reading}, and Tiny-Imagenet \cite{howard2017mobilenets}.  To validate LetheViT, we select various popular vision transformer models, including ViT-T/S/B, DeiT-T/S/B, and Swin-T/S. For faster convergence and better overall performance, we use the pre-trained models.

\noindent\textbf{Baselines.} To evaluate our method comprehensively, we select multiple baselines, including: (1)\textbf{ Retrain}: The most effective but computationally expensive method, retraining a model from scratch solely on the retained data. (2) \textbf{Fine-Tuning (FT)} \cite{warnecke2021machine,golatkar2020eternal}: A less intensive alternative requiring only minor adjustments to the original model via a few epochs on the retained data. (3)\textbf{ Gradient Ascent (GA)} \cite{graves2021amnesiac,thudi2022unrolling}: Updates the model parameters in the direction opposite to gradient descent, specifically using the forget dataset. (4) \textbf{Influence Unlearning (IU)} \cite{koh2017understanding,izzo2021approximate}: Estimates the impact of the forget set \( D_f \)
on model \( \mathcal{M}_0 \) using influence functions, then performs a Newton-step parameter update to negate it. (5)\textbf{Random Labels (RL)} \cite{golatkar2020eternal}: Trains on the full dataset after randomizing the labels of the forget set instances. (6) \textbf{$\ell_1$-sparse} \cite{liu2024model}: Induces weight sparsity through model pruning to achieve approximate unlearning. (7) \textbf{SalUn} \cite{fan2023salun}: Combines the RL approach with a gradient-based weight saliency map.

\noindent\textbf{Evaluation Metrics.} Aligning with prior works $\ell_1$-sparse  and SalUn, we adopt the following suite of evaluation metrics: \textbf{Forget Accuracy (FA)}: Measures model accuracy on the forget dataset post-unlearning. \textbf{Retain Accuracy (RA)}: Measures model accuracy on the retained dataset post-unlearning. Test Accuracy (TA): Measures model accuracy on a holdout test dataset, reflecting its generalization ability after unlearning. \textbf{Membership Inference Attack (MIA)} \cite{shokri2017membership}: A method assessing whether specific data points can be inferred as belonging to the original training set; used to detect residual information about supposedly forgotten data. Crucially, the ideal values for FA, RA, TA, and MIA are not simply maximizing or minimizing them; instead, they should exhibit minimal deviation from those achieved by the \textbf{Retrain} baseline (representing the unlearning gold standard) \cite{fan2023salun,liu2024model,tong2025robustmachineunlearningquantized}. To quantify overall performance, we introduce the \textbf{Average Gap (AG)}, computed as the mean absolute difference between each baseline method and the Retrain baseline across these four metrics after unlearning. A lower AG value indicates better unlearning efficacy, with zero representing ideal performance.

The implementation details and additional experiments (including those on more datasets and with different forgetting ratios) are included in the Appendix.

\begin{table}[t]
  \centering
  \small
  \setlength{\tabcolsep}{0.6mm}{
    \begin{tabular}{c|ccccc}
\toprule   
Ratio 
  & FA & RA & TA & MIA & AG$\color{blue}\downarrow$  \\
    \midrule
     Retrain &76.53&91.65&76.92&40.06 &0 \cellcolor{gray!20} \\
     \midrule
     5\% &80.90\textcolor{blue}{(4.37)} &94.09\textcolor{blue}{(2.44)} &75.04\textcolor{blue}{(1.88)} &37.60\textcolor{blue}{(2.46)} &2.79 \cellcolor{gray!20} \\
     10\% &81.47\textcolor{blue}{(4.94)} &94.55\textcolor{blue}{(2.90)} &75.52\textcolor{blue}{(1.40)} &36.45\textcolor{blue}{(3.61)} &3.21 \cellcolor{gray!20}  \\
     20\% &81.56\textcolor{blue}{(5.03)} &94.42\textcolor{blue}{(2.77)} &75.16\textcolor{blue}{(1.76)} &37.24\textcolor{blue}{(2.82)} &3.10 \cellcolor{gray!20} \\
     30\% &81.72\textcolor{blue}{(5.19)} &94.24\textcolor{blue}{(2.59)} &74.99\textcolor{blue}{(1.93)} &37.02\textcolor{blue}{(3.04)} &3.19 \cellcolor{gray!20}  \\
    \bottomrule
    \end{tabular}%
    }
    \caption{Performance of LetheViT under different masking ratios. The results are from DeiT-T on Tiny-Imagenet. }
  \label{tab3}
\end{table}

\subsection{Main Results}
Table \ref{tab2} compares the effectiveness of various Machine Unlearning (MU) methods on different Vision Transformer models (ViT-T/S/B, DeiT-T/S/B, Swin-T/S) using the Tiny-ImageNet dataset. The evaluation metrics include FA, RA, TA, MIA, and AG. LetheViT demonstrates significant advantages compared to existing methods such as FT, GA, IU, RL, and $\ell_1$-sparse, achieving the best forgetting effect. Specifically, in larger models like ViT-B and DeiT-B, LetheViT shows outstanding performance in the FA metric, reaching 86.03\% and 90.70\% respectively (with gaps of 1.59\% and 0.16\% compared to Retrain), showing the smallest gap with retraining. In terms of RA, LetheViT performs relatively stably. For example, in ViT-T, LetheViT achieves 91.55\% (4.22\% lower than Retrain), while SalUn only reaches 89.38\% (6.39\% lower than Retrain). This indicates that LetheViT can maintain the recognition ability for retained data while forgetting specific samples. In terms of TA, although LetheViT's gap is slightly higher than SalUn, its AG is significantly lower than all existing methods, achieving the best forgetting effect. More importantly, LetheViT achieves the smallest gap in Membership Inference Attack success rate (MIA) in each model series (ViT, DeiT, Swin), highlighting its excellent ability to suppress sensitive information leakage. For example, in ViT-B, LetheViT's MIA is 36.63\% (with a gap of 2.85\% compared to Retrain), while FT's MIA is 30.51\% (with a gap of 6.62\%), indicating that LetheViT is very effective in minimizing information leakage. LetheViT also performs well on the Swin series. For example, in the Swin-T model, LetheViT's AG is only 1.69\%, significantly better than SalUn's 3.98\% and $\ell_1$-sparse's 4.38\%. Its MIA is 25.12\% (with a gap of 1.24\% compared to Retrain), which is much better than $\ell_1$-sparse's 33.05\% (with a gap of 6.69\%). In the Swin-S model, LetheViT's AG is only 1.96\%, significantly better than SalUn's 4.06\% and $\ell_1$-sparse's 6.55\%. Its MIA is 19.74\% (with a gap of 1.34\% compared to Retrain), which is much better than SalUn's 35.00\% (with a gap of 13.92\%). Overall, LetheViT achieves the optimal forgetting effect on various Vision Transformer models by employing attention-guided contrastive learning to guide the model to forget specific samples while maintaining its recognition ability for the retained samples.

\subsection{Ablation Studies}

\noindent \textbf{Effect of Masking Ratio.} As shown in Table \ref{tab3}, we demonstrate the impact of different masking rates. When the masking ratio is 5\%, the average gap reaches its minimum value of 2.79\%, which is lower than those of other ratios (3.21\% for 10\%, 3.10\% for 20\%, and 3.19\% for 30\%). This indicates that 5\% is the optimal masking ratio, achieving the best forgetting effect. This finding is also consistent with the experimental results in Table \ref{tab1}: the 5\% masking ratio can preserve the model's recognition ability while reducing its memorization ability. However, higher ratios will disrupt the category outlines, leading to a decline in the forgetting effect.

\noindent \textbf{Effect of Masking Type.} In Table \ref{tab4}, we show the impact of different masking Types. Specifically, when using Zero masking, the model achieves a FA of 80.90\%, which is 4.37\% higher than that of Retrain. RA increases to 94.09\%, representing an improvement of 2.44\%. TA, however, decreases slightly to 75.04\%, a drop of 1.88\%. Meanwhile, the MIA is reduced to 37.60\%. The AG remains at a low level of 2.79\%. In contrast, Gaussian masking improves  FA to 82.27\% and  RA to 94.64\%. However, TA slightly drops to 75.24\% and the MIA is even lower at 36.27\%. Meanwhile, the AG increases to 3.55\%. This indicates that Gaussian masking introduces more noise compared to Zero masking, thereby reducing the forgetting effect.

\begin{table}[t]
  \centering
  \small
  \setlength{\tabcolsep}{0.12mm}{
    \begin{tabular}{c|ccccc}
\toprule   
Type 
  & FA & RA & TA & MIA & AG$\color{blue}\downarrow$  \\
    \midrule
     Retrain &76.53&91.65&76.92&40.06 &0 \cellcolor{gray!20} \\
     \midrule
     Zero &80.90\textcolor{blue}{(4.37)} &94.09\textcolor{blue}{(2.44)} &75.04\textcolor{blue}{(1.88)} &37.60\textcolor{blue}{(2.46)} &  2.79 \cellcolor{gray!20} \\
     Gaussian &82.27\textcolor{blue}{(5.74)} &94.64\textcolor{blue}{(2.99)}  &75.24\textcolor{blue}{(1.68)}  &36.27\textcolor{blue}{(3.79)}  &3.55 \cellcolor{gray!20}
  \\
    \bottomrule
    \end{tabular}%
    }
    \caption{Performance of LetheViT under different masking
types. The results are from DeiT-T on Tiny-Imagenet. }
  \label{tab4}
\end{table}

\begin{figure}[t]
\centering
\includegraphics[width=1\columnwidth]{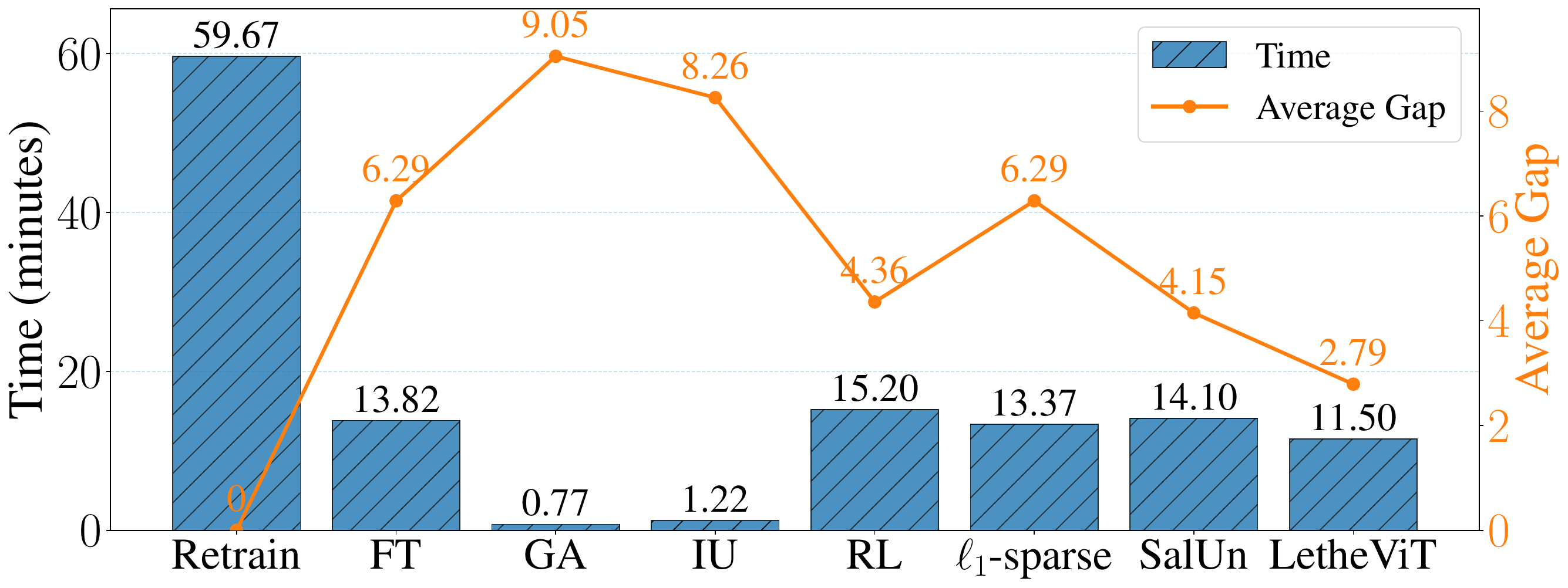} 
\caption{Efficiency and performance comparison. The results are from DeiT-T on Tiny-Imagenet.}
\label{fig3}
\end{figure}

\noindent \textbf{Efficiency Analysis.} Figure \ref{fig3} shows the efficiency of different methods. Specifically, the Retrain method requires 59.67 minutes, which is significantly longer than the Approximate Unlearning methods. Among the Approximate Unlearning methods, GA and IU achieve the highest efficiency. However, their average gaps are 9.05\% and 8.26\%, respectively, which are much higher than that of the Retrain method. This indicates that their forgetting effects are not satisfactory. In contrast, methods such as RL, FT, and SalUn improve the forgetting effect but at the cost of significantly increased time overhead. Compared to these methods, LetheViT not only achieves the best forgetting effect but also maintains a relatively low time cost.

\section{Related Work}

\noindent\textbf{Vision Transformer.} The Transformer \cite{vaswani2017attention} architecture, initially popular in natural language processing, has become dominant in computer vision. Unlike CNNs, it captures long-range visual relationships via self-attention. Vision Transformers (ViTs) \cite{dosovitskiy2020image,tong2025dfq} divide images into 16×16 patches as tokens, with a unique class token for classification. DeiT \cite{touvron2021training} enhances ViT's practicality through efficient training with limited data using knowledge distillation and data augmentation. Swin Transformer ~\cite{liu2021swin} uses a sliding window and hierarchical structure to capture local and global features efficiently while reducing computation. As ViTs become foundational in computer vision, privacy protection for ViTs is an important research direction. This paper focuses on  privacy protection of ViTs.

\noindent\textbf{Machine Unlearning.} Machine unlearning \cite{ginart2019making, bourtoule2021machine, sekhari2021remember,golatkar2020eternal} can remove the impact of specific samples on a model to protect privacy. The most effective method is to retrain the model from scratch using the retained dataset after removing the data points \cite{fan2023salun}, but this is computationally expensive, especially for large models. Thus, researchers are developing approximate unlearning methods \cite{tong2025robustmachineunlearningquantized,liu2024model}  to reduce costs while maintaining model performance after unlearning. However, existing methods such as
SalUn have not fully considered the characteristics of ViTs, especially their attention mechanisms. Meanwhile, recent works specifically targeting ViTs, such as NOVO \cite{roy2025novo} and Low-rank \cite{poppi2024unlearning} which focus on class-wise forgetting, and FRAMU \cite{shaik2024framu} which targets federated unlearning scenarios, are not directly applicable to random data forgetting. Unlike these methods, we propose an attention-based forgetting method specifically designed for the random forgetting scenario.

\noindent\textbf{Contrastive Learning.} Contrastive learning brings similar samples closer and pushes dissimilar ones farther to capture data structure and features. For example, SimCLR \cite{chen2020simple} constructs positive pairs from augmented images and optimizes consistency via a contrastive loss. SupCon \cite{khosla2020supervised} extends this to supervised settings using labels for intra-class compactness and inter-class separability. MoCo \cite{he2020momentum} uses a momentum-based queue to scale negatives and maintain consistency with a key encoder. These methods excel at representation learning but cannot be directly adapted for machine unlearning in ViTs. To address this, we propose a novel method using contrastive learning for unlearning specific samples while preserving model performance.

\section{Conclusions}
In this paper, we propose LetheViT, a Machine Unlearning method for ViTs. To achieve the forgetting of specific samples, we first explored the impact of masked images on the recognition and memory capabilities of ViT and found that zeroing out the patch with the highest attention score in the image and then performing inference does not degrade ViT's recognition ability, but weakens its memory of that image. Based on the above insights, we propose a contrastive Unlearning method for ViTs. Specifically, we input the masked image to generate positive logits and the original image to generate negative logits, guiding the model to forget specific details while preserving the general category outlines. The experiments demonstrate that LetheViT can achieve better forgetting effects than existing methods.

\bibliography{main}

\clearpage

\appendix

\begin{table*}[htpb]
  \centering
  \begin{tabular}{c|ccccc}
\toprule
\multirow{2}{*}{Method}  & \multicolumn{5}{c}{CIFAR-10 (ViT-T)}  \\
\cmidrule{2-6}
  & FA & RA & TA & MIA & AG$\color{blue}\downarrow$ \\
\midrule
\multicolumn{6}{c}{\cellcolor{gray!20}\textit{The proportion of forgotten data samples to all samples is 10\%}} \\
\midrule
 Retrain &96.87&99.91&97.59&6.11&\cellcolor{gray!20}0 \\
   FT &99.44\textcolor{blue}{(2.57)}&99.94\textcolor{blue}{(0.03)}&97.70\textcolor{blue}{(0.11)}&1.93\textcolor{blue}{(4.18)}& \cellcolor{gray!20}1.72 \\
    GA &99.46\textcolor{blue}{(2.59)}&99.58\textcolor{blue}{(0.33)}&97.63\textcolor{blue}{(0.04)}&1.91\textcolor{blue}{(4.20)}& \cellcolor{gray!20}1.79 \\
    IU &98.76\textcolor{blue}{(1.89)}&99.21\textcolor{blue}{(0.70)}&97.12\textcolor{blue}{(0.47)}&2.80\textcolor{blue}{(3.31)}& \cellcolor{gray!20}1.59 \\
    RL &96.76\textcolor{blue}{(0.11)}&99.39\textcolor{blue}{(0.52)}&96.66\textcolor{blue}{(0.93)}&14.24\textcolor{blue}{(8.13)}& \cellcolor{gray!20}2.42 \\
    $\ell_1$-sparse &99.42\textcolor{blue}{(2.55)}&99.94\textcolor{blue}{(0.03)}&97.71\textcolor{blue}{(0.12)}&1.93\textcolor{blue}{(4.18)}& \cellcolor{gray!20}1.72 \\
    SalUn &96.53\textcolor{blue}{(0.34)}&99.18\textcolor{blue}{(0.73)}&96.49\textcolor{blue}{(1.10)}&15.13\textcolor{blue}{(9.02)}& \cellcolor{gray!20}2.80 \\
LetheViT &96.62\textcolor{blue}{(0.25)}&97.20\textcolor{blue}{(2.71)}&95.43\textcolor{blue}{(2.16)}&5.49\textcolor{blue}{(0.62)}& \cellcolor{gray!20}\textbf{1.44} \\
\midrule
\multirow{2}{*}{Method}  & \multicolumn{5}{c}{CIFAR-10 (ViT-T)}  \\
\cmidrule{2-6}
  & FA & RA & TA & MIA & AG$\color{blue}\downarrow$ \\
\midrule
\multicolumn{6}{c}{\cellcolor{gray!20}\textit{The proportion of forgotten data samples to all samples is 30\%}} \\
\midrule
Retrain &96.09&99.92&97.19&6.09&\cellcolor{gray!20}0 \\
FT &99.54\textcolor{blue}{(3.45)}&99.94\textcolor{blue}{(0.02)}&97.74\textcolor{blue}{(0.55)}&1.70\textcolor{blue}{(4.39)}& \cellcolor{gray!20}2.10 \\
GA &99.54\textcolor{blue}{(3.45)}&99.56\textcolor{blue}{(0.36)}&97.61\textcolor{blue}{(0.42)}&1.52\textcolor{blue}{(4.57)}& \cellcolor{gray!20}2.20 \\
IU &97.93\textcolor{blue}{(1.84)}&98.33\textcolor{blue}{(1.59)}&96.39\textcolor{blue}{(0.80)}&3.81\textcolor{blue}{(2.28)}& \cellcolor{gray!20}1.63 \\
RL &95.33\textcolor{blue}{(0.76)}&98.39\textcolor{blue}{(1.53)}&95.42\textcolor{blue}{(1.77)}&16.99\textcolor{blue}{(10.90)}& \cellcolor{gray!20}3.74 \\
$\ell_1$-sparse &99.53\textcolor{blue}{(3.44)}&99.94\textcolor{blue}{(0.02)}&97.64\textcolor{blue}{(0.55)}&1.73\textcolor{blue}{(4.36)}& \cellcolor{gray!20}2.09\\
SalUn &98.09\textcolor{blue}{(2.00)}&98.32\textcolor{blue}{(1.60)}&95.52\textcolor{blue}{(1.67)}&15.57\textcolor{blue}{(9.48)}& \cellcolor{gray!20}3.69 \\
LetheViT &96.99\textcolor{blue}{(0.90)}&97.33\textcolor{blue}{(2.59)}&95.57\textcolor{blue}{(1.62)}&4.87\textcolor{blue}{(1.22)}& \cellcolor{gray!20}\textbf{1.58} \\
\midrule
\multirow{2}{*}{Method}  & \multicolumn{5}{c}{CIFAR-10 (ViT-T)}  \\
\cmidrule{2-6}
  & FA & RA & TA & MIA & AG$\color{blue}\downarrow$ \\
\midrule
\multicolumn{6}{c}{\cellcolor{gray!20}\textit{The proportion of forgotten data samples to all samples is 50\%}} \\
\midrule
Retrain &95.67&99.87&96.86&7.04&\cellcolor{gray!20}0 \\
FT &99.56\textcolor{blue}{(3.89)}&99.96\textcolor{blue}{(0.09)}&97.67\textcolor{blue}{(0.81)}&1.72\textcolor{blue}{(5.32)}& \cellcolor{gray!20}2.53 \\
GA &99.45\textcolor{blue}{(3.78)}&99.52\textcolor{blue}{(0.35)}&97.58\textcolor{blue}{(0.72)}&1.74\textcolor{blue}{(5.30)}& \cellcolor{gray!20}2.54 \\
IU &97.96\textcolor{blue}{(2.29)}&98.34\textcolor{blue}{(1.53)}&95.83\textcolor{blue}{(1.03)}&3.99\textcolor{blue}{(3.05)}& \cellcolor{gray!20}1.98 \\
RL &94.25\textcolor{blue}{(1.42)}&97.02\textcolor{blue}{(2.85)}&93.64\textcolor{blue}{(3.22)}&16.82\textcolor{blue}{(9.78)}& \cellcolor{gray!20}4.32 \\
$\ell_1$-sparse &99.38\textcolor{blue}{(3.71)}&99.97\textcolor{blue}{(0.10)}&97.62\textcolor{blue}{(0.76)}&2.50\textcolor{blue}{(4.54)}& \cellcolor{gray!20}2.28 \\
SalUn &91.41\textcolor{blue}{(4.26)}&93.74\textcolor{blue}{(6.13)}&91.09\textcolor{blue}{(5.77)}&18.48\textcolor{blue}{(11.44)}& \cellcolor{gray!20}6.40 \\
LetheViT &95.53\textcolor{blue}{(0.14)}&95.83\textcolor{blue}{(4.04)}&94.17\textcolor{blue}{(2.69)}&6.78\textcolor{blue}{(0.26)}& \cellcolor{gray!20}\textbf{1.78} \\
\bottomrule
\end{tabular}
\caption{Performance of various MU methods for ViT-T on CIFAR-10. The unlearning scenarios include 10\%, 30\%, and 50\% forgetting rates. \textbf{Bold} indicates the best AG.}
\label{tab5}
\end{table*}

\section{Appendix}

The organization of the appendix is as follows:
\begin{itemize}
    \item Appendix A: Implementation Details ;
    \item Appendix B: ViT-T/S on CIFAR-10/CIFAR-100;
    \item Appendix C: DeiT-T/S on SVHN.
\end{itemize}

\subsection{A. Implementation Details}
We follow the experimental settings of SalUn and $\ell_1$-sparse for the baseline methods. All experiments are conducted using the SGD optimizer. For FT and RL, we train each model for 10 epochs with learning rates sampled from the range [1e-4, 1e-2]. For GA, we train for 5 epochs with learning rates between [1e-7, 1e-5]. For IU,we vary the parameter $\alpha$, which relates to the Woodfisher Hessian Inverse approximation, within the range [1,20]. For $\ell_1$-sparse,  we search for the optimal 
$\gamma$ value in the range [1e-6, 1e-4],and explore learning rates between [1e-5, 1e-3]. For SalUn , we train for 10 epochs with learning rates sampled from [1e-4, 1e-2] and sparsity ratios in the range [0.1, 0.9]. For LetheViT, we apply the SGD optimizer with a batch size of 128. For ViT-T/S/B, we train for 10 epochs with learning rates in the range [1e-5,1e-3]. For Swin-T/S/B, we train for 10 epochs with learning rates in the range [1e-5,1e-3]. For DeiT-T/S/B, we train for 10 epochs with learning rates in the range [1e-4,1e-2]. We set the masking ratio to 5\%, the number of training epochs for the forget set to 2, and the number of training epochs for the retain set to 8. All experiments are conducted on a single NVIDIA RTX 4090 GPU.


\subsection{B. ViT-T/S on CIFAR-10/CIFAR-100 Datasets}

We report the experimental results of ViT-T and ViT-S on CIFAR-10 and CIFAR-100 in Tables~\ref{tab5}, \ref{tab6}, \ref{tab7}, and \ref{tab8}.
For ViT-T on CIFAR-10, LetheViT shows average performance gaps from the Retrain model of 1.44\%, 1.58\%, and 1.78\% at different forgetting ratios. Compared with the best-performing baseline, these gaps are reduced by 0.15\%, 0.05\%, and 0.20\%, respectively.
On CIFAR-100 with ViT-T, LetheViT achieves either the best or second-best results. The average performance gaps from Retrain are 2.64\%, 2.13\%, and 2.09\%, which are 0.37\%, 1.07\%, and 2.60\% smaller than those of the previous state-of-the-art methods.
For ViT-S on CIFAR-100, LetheViT yields average gaps from Retrain of 0.66\%, 0.95\%, and 1.74\%, indicating strong performance across all forgetting ratios.

\subsection{C. DeiT-T/S on SVHN Dataset}

We summarize the results of DeiT-T and DeiT-S on the SVHN dataset in Tables~\ref{tab9} and \ref{tab10}.
LetheViT delivers either the best or second-best performance for DeiT-T.
For DeiT-S, LetheViT achieves average gaps from Retrain of 0.47\%, 0.68\%, and 1.12\%, consistently outperforming other baselines under varying forgetting settings.

\begin{table*}[h!]
  \centering
  \begin{tabular}{c|ccccc}
\toprule
\multirow{2}{*}{Method}  & \multicolumn{5}{c}{CIFAR-10 (ViT-S)}  \\
\cmidrule{2-6}
  & FA & RA & TA & MIA & AG$\color{blue}\downarrow$ \\
\midrule
\multicolumn{6}{c}{\cellcolor{gray!20}\textit{The proportion of forgotten data samples to all samples is 10\%}} \\
\midrule
Retrain &98.49&99.99&98.65&2.96&\cellcolor{gray!20}0 \\
  FT &98.40\textcolor{blue}{(0.09)}&99.99\textcolor{blue}{(0.00)}&98.53\textcolor{blue}{(0.12)}&3.20\textcolor{blue}{(0.24)}& \cellcolor{gray!20}\textbf{0.11} \\  
    GA &99.77\textcolor{blue}{(1.28)}&99.84\textcolor{blue}{(0.15)}&98.47\textcolor{blue}{(0.18)}&0.62\textcolor{blue}{(2.34)}& \cellcolor{gray!20}0.99 \\  
    IU &99.71\textcolor{blue}{(1.22)}&99.84\textcolor{blue}{(0.15)}&98.49\textcolor{blue}{(0.16)}&0.75\textcolor{blue}{(2.21)}& \cellcolor{gray!20}0.94 \\  
    RL &97.69\textcolor{blue}{(0.80)}&99.85\textcolor{blue}{(0.14)}&97.86\textcolor{blue}{(0.79)}&11.17\textcolor{blue}{(8.21)}& \cellcolor{gray!20}2.49 \\  
    $\ell_1$-sparse &98.56\textcolor{blue}{(0.07)}&99.99\textcolor{blue}{(0.00)}&98.38\textcolor{blue}{(0.27)}&2.84\textcolor{blue}{(0.12)}& \cellcolor{gray!20}0.12 \\  
    SalUn &97.64\textcolor{blue}{(0.85)}&99.79\textcolor{blue}{(0.20)}&97.59\textcolor{blue}{(1.06)}&8.04\textcolor{blue}{(5.08)}& \cellcolor{gray!20}1.80 \\ 
LetheViT &98.64\textcolor{blue}{(0.15)}&99.99\textcolor{blue}{(0.00)}&98.54\textcolor{blue}{(0.11)}&3.49\textcolor{blue}{(0.53)}& \cellcolor{gray!20}0.20 \\
\midrule
\multirow{2}{*}{Method}  & \multicolumn{5}{c}{CIFAR-10 (ViT-S)}  \\
\cmidrule{2-6}
  & FA & RA & TA & MIA & AG$\color{blue}\downarrow$ \\
\midrule
\multicolumn{6}{c}{\cellcolor{gray!20}\textit{The proportion of forgotten data samples to all samples is 30\%}} \\
\midrule
Retrain &98.40&99.98&98.54&3.01&\cellcolor{gray!20}0 \\
FT &98.82\textcolor{blue}{(0.42)}&99.99\textcolor{blue}{(0.01)}&98.34\textcolor{blue}{(0.20)}&2.89\textcolor{blue}{(0.12)}& \cellcolor{gray!20}\textbf{0.19} \\
GA &99.82\textcolor{blue}{(1.42)}&99.83\textcolor{blue}{(0.15)}&98.49\textcolor{blue}{(0.05)}&0.52\textcolor{blue}{(2.49)}& \cellcolor{gray!20}1.03 \\
IU &99.41\textcolor{blue}{(1.01)}&99.49\textcolor{blue}{(0.49)}&98.05\textcolor{blue}{(0.49)}&1.35\textcolor{blue}{(1.66)}& \cellcolor{gray!20}0.91 \\
RL &97.44\textcolor{blue}{(0.96)}&99.71\textcolor{blue}{(0.27)}&97.38\textcolor{blue}{(1.16)}&9.24\textcolor{blue}{(6.23)}& \cellcolor{gray!20}2.16 \\
$\ell_1$-sparse &98.76\textcolor{blue}{(0.36)}&99.98\textcolor{blue}{(0.00)}&98.38\textcolor{blue}{(0.16)}&2.91\textcolor{blue}{(0.10)}& \cellcolor{gray!20}0.16 \\
SalUn &97.24\textcolor{blue}{(1.16)}&99.31\textcolor{blue}{(0.67)}&96.87\textcolor{blue}{(1.67)}&10.07\textcolor{blue}{(7.06)}& \cellcolor{gray!20}2.64 \\
LetheViT &98.58\textcolor{blue}{(0.18)}&99.98\textcolor{blue}{(0.00)}&98.35\textcolor{blue}{(0.19)}&3.41\textcolor{blue}{(0.45)}& \cellcolor{gray!20}0.21 \\
\midrule
\multirow{2}{*}{Method}  & \multicolumn{5}{c}{CIFAR-10 (ViT-S)}  \\
\cmidrule{2-6}
  & FA & RA & TA & MIA & AG$\color{blue}\downarrow$ \\
\midrule
\multicolumn{6}{c}{\cellcolor{gray!20}\textit{The proportion of forgotten data samples to all samples is 50\%}} \\
\midrule
Retrain &98.24&99.97&98.17&3.69&\cellcolor{gray!20}0 \\
FT &98.76\textcolor{blue}{(0.52)}&99.98\textcolor{blue}{(0.01)}&98.06\textcolor{blue}{(0.11)}&3.17\textcolor{blue}{(0.52)}& \cellcolor{gray!20}0.29 \\
GA &99.80\textcolor{blue}{(1.56)}&99.83\textcolor{blue}{(0.14)}&98.49\textcolor{blue}{(0.32)}&0.57\textcolor{blue}{(3.12)}& \cellcolor{gray!20}1.29 \\
IU &99.15\textcolor{blue}{(0.91)}&99.28\textcolor{blue}{(0.69)}&97.79\textcolor{blue}{(0.38)}&1.79\textcolor{blue}{(1.90)}& \cellcolor{gray!20}0.97 \\
RL &96.92\textcolor{blue}{(1.32)}&99.35\textcolor{blue}{(0.62)}&96.45\textcolor{blue}{(1.72)}&10.16\textcolor{blue}{(6.47)}& \cellcolor{gray!20}2.53 \\
$\ell_1$-sparse &98.79\textcolor{blue}{(0.55)}&99.98\textcolor{blue}{(0.01)}&98.17\textcolor{blue}{(0.00)}&3.04\textcolor{blue}{(0.65)}& \cellcolor{gray!20}0.30 \\
SalUn &97.29\textcolor{blue}{(0.95)}&99.38\textcolor{blue}{(0.59)}&96.91\textcolor{blue}{(1.26)}&12.53\textcolor{blue}{(8.84)}& \cellcolor{gray!20}2.91 \\
LetheViT &98.44\textcolor{blue}{(0.20)}&99.99\textcolor{blue}{(0.02)}&98.04\textcolor{blue}{(0.13)}&4.12\textcolor{blue}{(0.43)}& \cellcolor{gray!20}\textbf{0.20} \\
\bottomrule
\end{tabular}
\caption{Performance of various MU methods for ViT-S on CIFAR-10. The unlearning scenarios include 10\%, 30\%, and 50\% forgetting rates. \textbf{Bold} indicates the best AG.}
\label{tab6}
\end{table*}

\begin{table*}[h!]
  \centering
  \begin{tabular}{c|ccccc}
\toprule
\multirow{2}{*}{Method}  & \multicolumn{5}{c}{CIFAR-100 (ViT-T)}  \\
\cmidrule{2-6}
  & FA & RA & TA & MIA & AG$\color{blue}\downarrow$ \\
\midrule
\multicolumn{6}{c}{\cellcolor{gray!20}\textit{The proportion of forgotten data samples to all samples is 10\%}} \\
\midrule
Retrain &85.82&97.01&84.71&21.42&\cellcolor{gray!20}0 \\
FT &83.24\textcolor{blue}{(2.58)}&99.60\textcolor{blue}{(2.59)}&80.83\textcolor{blue}{(3.88)}&31.88\textcolor{blue}{(10.46)}& \cellcolor{gray!20} 4.88 \\
GA &95.29\textcolor{blue}{(9.47)}&96.05\textcolor{blue}{(0.96)}&85.58\textcolor{blue}{(0.87)}&9.08\textcolor{blue}{(12.34)}& \cellcolor{gray!20} 5.91 \\
IU &96.09\textcolor{blue}{(10.27)}&96.90\textcolor{blue}{(0.11)}&86.33\textcolor{blue}{(1.62)}&8.80\textcolor{blue}{(12.62)}& \cellcolor{gray!20} 6.16 \\
RL &94.04\textcolor{blue}{(8.22)}&98.42\textcolor{blue}{(1.41)}&86.05\textcolor{blue}{(1.34)}&30.87\textcolor{blue}{(9.45)}& \cellcolor{gray!20} 5.11 \\
$\ell_1$-sparse &87.76\textcolor{blue}{(1.94)}&99.42\textcolor{blue}{(2.41)}&86.51\textcolor{blue}{(1.80)}&24.00\textcolor{blue}{(2.58)}& \cellcolor{gray!20} \textbf{2.18} \\
SalUn &94.67\textcolor{blue}{(8.85)}&97.90\textcolor{blue}{(0.89)}&85.70\textcolor{blue}{(0.99)}&22.73\textcolor{blue}{(1.31)}& \cellcolor{gray!20} 3.01 \\
LetheViT &89.75\textcolor{blue}{(3.93)}&99.75\textcolor{blue}{(2.74)}&86.36\textcolor{blue}{(1.65)}&23.67\textcolor{blue}{(2.25)}&\cellcolor{gray!20}2.64 \\
\midrule
\multirow{2}[4]{*}{Method}  & \multicolumn{5}{c}{CIFAR-100 (ViT-T)} \\
\cmidrule{2-6}
   & FA & RA & TA & MIA & AG$\color{blue}\downarrow$ \\
    \midrule
\multicolumn{6}{c}{\cellcolor{gray!20}\textit{The proportion of forgotten data samples to all samples is 30\%}} \\
\midrule
Retrain &82.93  &96.17  &83.12  &25.99  & \cellcolor{gray!20} 0 \\
FT &79.88\textcolor{blue}{(3.05)} &99.66\textcolor{blue}{(3.49)} &79.51\textcolor{blue}{(3.61)} &35.25\textcolor{blue}{(9.26)} & \cellcolor{gray!20} 4.85 \\
GA &97.44\textcolor{blue}{(14.51)} &97.66\textcolor{blue}{(1.49)} &87.10\textcolor{blue}{(3.98)} &7.93\textcolor{blue}{(18.06)} & \cellcolor{gray!20} 9.51 \\
IU &93.45\textcolor{blue}{(10.52)} &94.90\textcolor{blue}{(1.27)} &84.56\textcolor{blue}{(1.44)} &10.90\textcolor{blue}{(15.09)} & \cellcolor{gray!20} 7.08 \\
RL &93.97\textcolor{blue}{(11.04)} &97.19\textcolor{blue}{(1.02)} &85.10\textcolor{blue}{(1.98)} &30.50\textcolor{blue}{(4.51)} & \cellcolor{gray!20} 4.64 \\
$\ell_1$-sparse &93.67\textcolor{blue}{(10.74)} &99.93\textcolor{blue}{(3.76)} &87.46\textcolor{blue}{(4.34)} &19.00\textcolor{blue}{(6.99)} & \cellcolor{gray!20} 6.46 \\
SalUn &93.90\textcolor{blue}{(10.97)} &96.34\textcolor{blue}{(0.17)} &84.76\textcolor{blue}{(1.64)} &25.96\textcolor{blue}{(0.03)} & \cellcolor{gray!20} 3.20 \\
LetheViT &86.65\textcolor{blue}{(0.83)}&99.70\textcolor{blue}{(3.57)}&84.91\textcolor{blue}{(1.79)}&28.33\textcolor{blue}{(2.34)}&       \cellcolor{gray!20} \textbf{2.13} \\
\midrule
\multirow{2}[4]{*}{Method}  & \multicolumn{5}{c}{CIFAR-100 (ViT-T)} \\
\cmidrule{2-6}
   & FA & RA & TA & MIA & AG$\color{blue}\downarrow$ \\
    \midrule
\multicolumn{6}{c}{\cellcolor{gray!20}\textit{The proportion of forgotten data samples to all samples is 50\%}} \\
\midrule
Retrain &80.96  &95.37  &81.16  &29.85  & \cellcolor{gray!20} 0 \\
FT &78.92\textcolor{blue}{(2.04)} &99.87\textcolor{blue}{(4.50)} &78.40\textcolor{blue}{(2.76)} &38.49\textcolor{blue}{(8.64)} & \cellcolor{gray!20} 4.48 \\
GA &97.53\textcolor{blue}{(16.57)} &97.57\textcolor{blue}{(2.20)} &87.00\textcolor{blue}{(5.84)} &8.37\textcolor{blue}{(21.48)} & \cellcolor{gray!20} 11.52 \\
IU &88.16\textcolor{blue}{(7.20)} &89.88\textcolor{blue}{(5.49)} &80.41\textcolor{blue}{(0.75)} &15.70\textcolor{blue}{(14.15)} & \cellcolor{gray!20} 6.90 \\
RL &93.11\textcolor{blue}{(12.15)} &95.52\textcolor{blue}{(0.15)} &84.45\textcolor{blue}{(3.29)} &29.61\textcolor{blue}{(0.24)} & \cellcolor{gray!20} 3.96 \\
$\ell_1$-sparse &93.95\textcolor{blue}{(12.99)} &99.93\textcolor{blue}{(4.56)} &87.36\textcolor{blue}{(6.20)} &20.18\textcolor{blue}{(9.67)} & \cellcolor{gray!20} 8.36 \\
SalUn &92.36\textcolor{blue}{(11.40)} &94.27\textcolor{blue}{(1.10)} &83.60\textcolor{blue}{(2.44)} &26.03\textcolor{blue}{(3.82)} & \cellcolor{gray!20} 4.69 \\
LetheViT &81.84\textcolor{blue}{(0.88)}&99.23\textcolor{blue}{(3.86)}&80.89\textcolor{blue}{(0.27)}&33.21\textcolor{blue}{(3.36)}& \cellcolor{gray!20} \textbf{2.09} \\
\bottomrule
\end{tabular}
\caption{Performance of various MU methods for ViT-T on CIFAR-100. The unlearning scenarios include 10\%, 30\%, and 50\% forgetting rates. \textbf{Bold} indicates the best AG.}
\label{tab7}
\end{table*}

\begin{table*}[h!]
  \centering
  \begin{tabular}{c|ccccc}
\toprule
\multirow{2}{*}{Method}  & \multicolumn{5}{c}{CIFAR-100 (ViT-S)}  \\
\cmidrule{2-6}
  & FA & RA & TA & MIA & AG$\color{blue}\downarrow$ \\
\midrule
\multicolumn{6}{c}{\cellcolor{gray!20}\textit{The proportion of forgotten data samples to all samples is 10\%}} \\
\midrule
 Retrain &92.02&99.75&90.99&15.47&\cellcolor{gray!20}0 \\
  FT &90.73\textcolor{blue}{(1.29)}&99.92\textcolor{blue}{(0.17)}&89.53\textcolor{blue}{(1.46)}&19.98\textcolor{blue}{(4.51)}& \cellcolor{gray!20}1.86 \\  
    GA &98.67\textcolor{blue}{(6.65)}&98.41\textcolor{blue}{(1.34)}&91.10\textcolor{blue}{(0.11)}&5.27\textcolor{blue}{(10.20)}& \cellcolor{gray!20}4.58 \\  
    IU &97.82\textcolor{blue}{(5.80)}&98.16\textcolor{blue}{(1.59)}&90.85\textcolor{blue}{(0.14)}&6.58\textcolor{blue}{(8.89)}& \cellcolor{gray!20}4.11 \\  
    RL &91.49\textcolor{blue}{(0.53)}&99.49\textcolor{blue}{(0.26)}&90.48\textcolor{blue}{(0.51)}&33.96\textcolor{blue}{(18.49)}& \cellcolor{gray!20}4.95 \\  
    $\ell_1$-sparse &89.71\textcolor{blue}{(2.31)}&99.90\textcolor{blue}{(0.15)}&88.59\textcolor{blue}{(2.40)}&21.31\textcolor{blue}{(5.84)}& \cellcolor{gray!20}2.68 \\  
    SalUn &97.95\textcolor{blue}{(5.93)}&98.07\textcolor{blue}{(1.68)}&90.05\textcolor{blue}{(0.94)}&9.04\textcolor{blue}{(6.43)}& \cellcolor{gray!20}3.75 \\  
LetheViT  &92.38 \textcolor{blue}{(0.36)} &99.91 \textcolor{blue}{(0.16)} &90.31 \textcolor{blue}{(0.68)} &16.91 \textcolor{blue}{(1.44)} & \cellcolor{gray!20}\textbf{0.66} \\
\midrule
\multirow{2}{*}{Method}  & \multicolumn{5}{c}{CIFAR-100 (ViT-S)}  \\
\cmidrule{2-6}
  & FA & RA & TA & MIA & AG$\color{blue}\downarrow$ \\
\midrule
\multicolumn{6}{c}{\cellcolor{gray!20}\textit{The proportion of forgotten data samples to all samples is 30\%}} \\
\midrule
Retrain &90.26  &99.76  &90.40  &18.27  & \cellcolor{gray!20}0 \\
FT      &94.08 \textcolor{blue}{(3.82)} &99.95 \textcolor{blue}{(0.19)} &90.58 \textcolor{blue}{(0.18)} &15.68 \textcolor{blue}{(2.59)} & \cellcolor{gray!20}\textbf{1.70} \\
GA      &98.27 \textcolor{blue}{(8.01)} &98.50 \textcolor{blue}{(1.26)} &91.09 \textcolor{blue}{(0.69)} &5.57 \textcolor{blue}{(12.70)} & \cellcolor{gray!20}5.66 \\
IU      &97.14 \textcolor{blue}{(6.88)} &97.66 \textcolor{blue}{(2.10)} &89.99 \textcolor{blue}{(0.41)} &7.36 \textcolor{blue}{(10.91)} & \cellcolor{gray!20}5.08 \\
RL      &94.32 \textcolor{blue}{(4.06)} &99.23 \textcolor{blue}{(0.53)} &89.80 \textcolor{blue}{(0.60)} &36.46 \textcolor{blue}{(18.19)} & \cellcolor{gray!20}5.85 \\
$\ell_1$-sparse &94.24 \textcolor{blue}{(3.98)} &99.97\textcolor{blue}{(0.21)} &90.72 \textcolor{blue}{(0.32)} &15.42 \textcolor{blue}{(2.85)} & \cellcolor{gray!20}1.84 \\
SalUn   &97.75 \textcolor{blue}{(7.49)} &98.04 \textcolor{blue}{(1.72)} &90.41 \textcolor{blue}{(0.01)} &10.96 \textcolor{blue}{(7.31)} & \cellcolor{gray!20}4.13 \\
LetheViT  &90.95 \textcolor{blue}{0.69} &99.90 \textcolor{blue}{(0.14)} &89.49 \textcolor{blue}{(0.91)} &20.31 \textcolor{blue}{(2.04)} & \cellcolor{gray!20}0.95 \\
\midrule
\multirow{2}{*}{Method}  & \multicolumn{5}{c}{CIFAR-100 (ViT-S)}  \\
\cmidrule{2-6}
  & FA & RA & TA & MIA & AG$\color{blue}\downarrow$ \\
\midrule
\multicolumn{6}{c}{\cellcolor{gray!20}\textit{The proportion of forgotten data samples to all samples is 50\%}} \\
\midrule
Retrain &89.68  &99.68  &89.69  &20.93  & \cellcolor{gray!20}0 \\
FT      &94.17 \textcolor{blue}{(4.49)} &99.95 \textcolor{blue}{(0.27)} &90.49 \textcolor{blue}{(0.80)} &15.95\textcolor{blue}{(4.98)} & \cellcolor{gray!20}\textbf{2.64} \\
GA      &98.41 \textcolor{blue}{(8.73)} &98.45 \textcolor{blue}{(1.23)} &91.09 \textcolor{blue}{(1.40)} &5.82 \textcolor{blue}{(15.11)} & \cellcolor{gray!20}6.62 \\
IU      &96.32 \textcolor{blue}{(6.64)} &97.12 \textcolor{blue}{(2.56)} &89.51 \textcolor{blue}{(0.18)} &7.67 \textcolor{blue}{(13.26)} & \cellcolor{gray!20}5.66 \\
RL      &96.19 \textcolor{blue}{(6.51)} &98.74 \textcolor{blue}{(0.94)} &90.06\textcolor{blue}{(0.37)} &44.92 \textcolor{blue}{(23.99)} & \cellcolor{gray!20}7.95 \\
$\ell_1$-sparse &94.31 \textcolor{blue}{(4.63)} &99.96 \textcolor{blue}{(0.28)} &90.47 \textcolor{blue}{(0.78)} &15.99 \textcolor{blue}{(4.94)} & \cellcolor{gray!20}2.66 \\
SalUn   &90.33 \textcolor{blue}{(0.65)} &91.00 \textcolor{blue}{(8.68)} &83.63 \textcolor{blue}{(6.06)} &24.38 \textcolor{blue}{(3.45)} & \cellcolor{gray!20}4.71 \\
LetheViT  &88.65 \textcolor{blue}{(1.03)} &99.83 \textcolor{blue}{(0.15)} &87.74 \textcolor{blue}{(1.95)} &24.76 \textcolor{blue}{(3.83)} & \cellcolor{gray!20}1.74 \\
\bottomrule
\end{tabular}
\caption{Performance of various MU methods for ViT-S on CIFAR-100. The unlearning scenarios include 10\%, 30\%, and 50\% forgetting rates. \textbf{Bold} indicates the best AG.}
\label{tab8}
\end{table*}

\begin{table*}[h!]
  \centering
  \begin{tabular}{c|ccccc}
\toprule
\multirow{2}{*}{Method}  & \multicolumn{5}{c}{SVHN (DeiT-T)}  \\
\cmidrule{2-6}
  & FA & RA & TA & MIA & AG$\color{blue}\downarrow$ \\
\midrule
\multicolumn{6}{c}{\cellcolor{gray!20}\textit{The proportion of forgotten data samples to all samples is 10\%}} \\
\midrule
 Retrain &95.37&98.54&95.86&7.25&\cellcolor{gray!20}0 \\
  FT &96.29\textcolor{blue}{(0.92)}&99.83\textcolor{blue}{(1.29)}&97.21\textcolor{blue}{(1.35)}&6.96\textcolor{blue}{(0.29)}& \cellcolor{gray!20}\textbf{0.96} \\  
    GA &97.66\textcolor{blue}{(2.29)}&97.66\textcolor{blue}{(0.88)}&95.93\textcolor{blue}{(0.07)}&5.98\textcolor{blue}{(1.27)}& \cellcolor{gray!20}1.13 \\  
    IU &96.71\textcolor{blue}{(1.34)}&97.12\textcolor{blue}{(1.42)}&95.61\textcolor{blue}{(0.25)}&8.21\textcolor{blue}{(0.96)}& \cellcolor{gray!20}0.99 \\  
    RL &95.06\textcolor{blue}{(0.31)}&98.47\textcolor{blue}{(0.07)}&95.37\textcolor{blue}{(0.49)}&15.64\textcolor{blue}{(8.39)}& \cellcolor{gray!20}2.32 \\  
    $\ell_1$-sparse &96.25\textcolor{blue}{(0.88)}&99.62\textcolor{blue}{(1.08)}&96.52\textcolor{blue}{(0.66)}&6.58\textcolor{blue}{(0.67)}& \cellcolor{gray!20}0.82 \\ 
    SalUn &93.10\textcolor{blue}{(2.27)}&96.22\textcolor{blue}{(2.32)}&93.93\textcolor{blue}{(1.93)}&13.04\textcolor{blue}{(5.79)}& \cellcolor{gray!20}3.08 \\ 
LetheViT  &96.28 \textcolor{blue}{(0.91)} &99.20 \textcolor{blue}{(0.66)} &96.55 \textcolor{blue}{(0.69)} &7.04 \textcolor{blue}{(0.21)} &\cellcolor{gray!20}0.62 \\
\midrule
\multirow{2}{*}{Method}  & \multicolumn{5}{c}{SVHN (DeiT-T)}  \\
\cmidrule{2-6}
  & FA & RA & TA & MIA & AG$\color{blue}\downarrow$ \\
\midrule
\multicolumn{6}{c}{\cellcolor{gray!20}\textit{The proportion of forgotten data samples to all samples is 30\%}} \\
\midrule
Retrain &94.23  &98.38  &95.38  &9.33  & \cellcolor{gray!20}0 \\
FT      &96.18 \textcolor{blue}{(1.95)} &99.66 \textcolor{blue}{(1.28)} &96.29 \textcolor{blue}{(0.91)} &7.46 \textcolor{blue}{(1.87)} & \cellcolor{gray!20}\textbf{1.50} \\
GA      &97.19 \textcolor{blue}{(2.96)} &97.39 \textcolor{blue}{(0.99)} &95.62 \textcolor{blue}{(0.24)} &7.35 \textcolor{blue}{(1.98)} & \cellcolor{gray!20}1.54 \\
IU      &92.80 \textcolor{blue}{(1.43)} &93.38 \textcolor{blue}{(5.00)} &92.36 \textcolor{blue}{(3.02)} &14.91 \textcolor{blue}{(5.58)} & \cellcolor{gray!20}3.76 \\
RL      &93.67 \textcolor{blue}{(0.56)} &97.03 \textcolor{blue}{(1.35)} &93.56 \textcolor{blue}{(1.82)} &15.47\textcolor{blue}{(6.14)} & \cellcolor{gray!20}2.47 \\
$\ell_1$-sparse &96.16 \textcolor{blue}{(1.93)} &99.66 \textcolor{blue}{(1.28)} &96.33 \textcolor{blue}{(0.95)} &7.59 \textcolor{blue}{(1.74)} & \cellcolor{gray!20}1.48 \\
SalUn   &93.42 \textcolor{blue}{(0.81)} &96.14 \textcolor{blue}{(2.24)} &93.15 \textcolor{blue}{(2.23)} &17.04 \textcolor{blue}{(7.71)} & \cellcolor{gray!20}3.25 \\
LetheViT  &96.12 \textcolor{blue}{(1.89))} &99.16 \textcolor{blue}{(0.78)} &96.57 \textcolor{blue}{(1.19)} &7.56 \textcolor{blue}{(1.77)} & \cellcolor{gray!20}1.40\\
\midrule
\multirow{2}{*}{Method}  & \multicolumn{5}{c}{SVHN (DeiT-T)}  \\
\cmidrule{2-6}
  & FA & RA & TA & MIA & AG$\color{blue}\downarrow$ \\
\midrule
\multicolumn{6}{c}{\cellcolor{gray!20}\textit{The proportion of forgotten data samples to all samples is 50\%}} \\
\midrule
Retrain &93.79  &97.93  &94.58  &11.43  & \cellcolor{gray!20}0 \\
FT      &96.10 \textcolor{blue}{(2.31)} &99.69 \textcolor{blue}{(1.76)} &96.20 \textcolor{blue}{(1.62)} &7.60 \textcolor{blue}{(3.83)} & \cellcolor{gray!20}\textbf{2.38} \\
GA      &96.45 \textcolor{blue}{(2.66)} &96.46 \textcolor{blue}{(1.47)} &94.91 \textcolor{blue}{(0.33)} &8.37\textcolor{blue}{(3.06)} & \cellcolor{gray!20}1.88 \\
IU      &90.00 \textcolor{blue}{(3.79)} &90.38 \textcolor{blue}{(7.55)} &89.68 \textcolor{blue}{(4.90)} &19.72 \textcolor{blue}{(8.29)} & \cellcolor{gray!20}6.14 \\
RL      &91.18 \textcolor{blue}{(2.61)} &93.05 \textcolor{blue}{(4.88)} &90.81 \textcolor{blue}{(3.77)} &23.40\textcolor{blue}{(11.97)} & \cellcolor{gray!20}5.81 \\
$\ell_1$-sparse &96.11 \textcolor{blue}{(2.32)} &99.69 \textcolor{blue}{(1.76)} &96.21 \textcolor{blue}{(1.63)} &7.69 \textcolor{blue}{(3.74)} & \cellcolor{gray!20}2.36 \\
SalUn   &83.13 \textcolor{blue}{(10.66)} &86.09 \textcolor{blue}{(11.84)} &82.41 \textcolor{blue}{(12.17)} &31.38 \textcolor{blue}{(19.95)} & \cellcolor{gray!20}13.66 \\
LetheViT  &96.00 \textcolor{blue}{(2.21)} &99.25 \textcolor{blue}{(1.32)} &96.48 \textcolor{blue}{(1.90)} &7.81 \textcolor{blue}{(3.62))} & \cellcolor{gray!20}2.26 \\
\bottomrule
\end{tabular}
\caption{Performance of various MU methods for DeiT-T on SVHN. The unlearning scenarios include 10\%, 30\%, and 50\% forgetting rates. \textbf{Bold} indicates the best AG.}
\label{tab9}
\end{table*}

\begin{table*}[h!]
  \centering
  \begin{tabular}{c|ccccc}
\toprule
\multirow{2}{*}{Method}  & \multicolumn{5}{c}{SVHN (DeiT-S)}  \\
\cmidrule{2-6}
  & FA & RA & TA & MIA & AG$\color{blue}\downarrow$ \\
\midrule
\multicolumn{6}{c}{\cellcolor{gray!20}\textit{The proportion of forgotten data samples to all samples is 10\%}} \\
\midrule
Retrain &95.57&99.64&96.54&7.28&\cellcolor{gray!20}0 \\
FT &96.50\textcolor{blue}{(0.93)}&99.86\textcolor{blue}{(0.22)}&97.09\textcolor{blue}{(0.55)}&6.69\textcolor{blue}{(0.59)}& \cellcolor{gray!20}\textbf{0.57} \\
GA &97.97\textcolor{blue}{(2.40)}&97.91\textcolor{blue}{(1.73)}&96.00\textcolor{blue}{(0.54)}&5.52\textcolor{blue}{(1.76)}& \cellcolor{gray!20}1.61 \\
IU &97.44\textcolor{blue}{(1.87)}&97.96\textcolor{blue}{(1.68)}&95.94\textcolor{blue}{(0.60)}&6.22\textcolor{blue}{(1.06)}& \cellcolor{gray!20}1.30 \\
RL &96.72\textcolor{blue}{(1.15)}&99.77\textcolor{blue}{(0.13)}&97.34\textcolor{blue}{(0.80)}&12.88\textcolor{blue}{(5.60)}& \cellcolor{gray!20}1.92 \\
$\ell_1$-sparse &96.41\textcolor{blue}{(0.84)}&99.87\textcolor{blue}{(0.23)}&97.00\textcolor{blue}{(0.46)}&7.04\textcolor{blue}{(0.24)}& \cellcolor{gray!20}0.44 \\
SalUn &97.06\textcolor{blue}{(1.49)}&99.54\textcolor{blue}{(0.10)}&97.26\textcolor{blue}{(0.72)}&12.26\textcolor{blue}{(4.98)}& \cellcolor{gray!20}1.82 \\
LetheViT &96.51\textcolor{blue}{(0.94)}&99.66\textcolor{blue}{(0.02)}&96.96\textcolor{blue}{(0.42)}&6.80\textcolor{blue}{(0.48)}& \cellcolor{gray!20}0.47 \\
\midrule
\multirow{2}{*}{Method}  & \multicolumn{5}{c}{SVHN (DeiT-S)}  \\
\cmidrule{2-6}
  & FA & RA & TA & MIA & AG$\color{blue}\downarrow$ \\
\midrule
\multicolumn{6}{c}{\cellcolor{gray!20}\textit{The proportion of forgotten data samples to all samples is 30\%}} \\
\midrule
Retrain &95.10&99.58&96.14&8.30&\cellcolor{gray!20}0 \\
FT &96.08\textcolor{blue}{(0.98)}&99.86\textcolor{blue}{(0.28)}&96.91\textcolor{blue}{(0.77)}&7.72\textcolor{blue}{(0.58)}& \cellcolor{gray!20}\textbf{0.65} \\
GA &95.78\textcolor{blue}{(0.68)}&96.00\textcolor{blue}{(3.58)}&94.38\textcolor{blue}{(1.76)}&9.09\textcolor{blue}{(0.79)}& \cellcolor{gray!20}1.70 \\
IU &93.95\textcolor{blue}{(1.15)}&94.46\textcolor{blue}{(5.12)}&93.68\textcolor{blue}{(2.46)}&12.64\textcolor{blue}{(4.34)}& \cellcolor{gray!20}3.27 \\
RL &96.14\textcolor{blue}{(1.04)}&99.31\textcolor{blue}{(0.27)}&96.18\textcolor{blue}{(0.04)}&11.31\textcolor{blue}{(3.01)}& \cellcolor{gray!20}1.09 \\
$\ell_1$-sparse &95.98\textcolor{blue}{(0.88)}&99.89\textcolor{blue}{(0.31)}&96.97\textcolor{blue}{(0.83)}&7.41\textcolor{blue}{(0.89)}& \cellcolor{gray!20}0.75 \\
SalUn &96.44\textcolor{blue}{(1.34)}&99.17\textcolor{blue}{(0.41)}&96.46\textcolor{blue}{(0.32)}&10.69\textcolor{blue}{(2.39)}& \cellcolor{gray!20}1.12 \\
LetheViT &96.01\textcolor{blue}{(0.91)}&99.63\textcolor{blue}{(0.05)}&96.88\textcolor{blue}{(0.74)}&7.30\textcolor{blue}{(1.00)}& \cellcolor{gray!20}0.68 \\

\midrule
\multirow{2}{*}{Method}  & \multicolumn{5}{c}{SVHN (DeiT-S)}  \\
\cmidrule{2-6}
  & FA & RA & TA & MIA & AG$\color{blue}\downarrow$ \\
\midrule
\multicolumn{6}{c}{\cellcolor{gray!20}\textit{The proportion of forgotten data samples to all samples is 50\%}} \\
\midrule
Retrain &94.38&99.55&95.61&9.23&\cellcolor{gray!20}0 \\
FT &96.03\textcolor{blue}{(1.65)}&99.89\textcolor{blue}{(0.34)}&96.64\textcolor{blue}{(1.03)}&7.66\textcolor{blue}{(1.57)}& \cellcolor{gray!20}\textbf{1.15} \\
GA &93.03\textcolor{blue}{(1.35)}&98.05\textcolor{blue}{(1.50)}&93.10\textcolor{blue}{(2.51)}&11.56\textcolor{blue}{(2.33)}& \cellcolor{gray!20}1.92 \\
IU &90.35\textcolor{blue}{(4.03)}&90.93\textcolor{blue}{(8.62)}&90.61\textcolor{blue}{(5.00)}&17.11\textcolor{blue}{(7.88)}& \cellcolor{gray!20}6.38 \\
RL &93.47\textcolor{blue}{(0.91)}&97.57\textcolor{blue}{(1.98)}&93.51\textcolor{blue}{(2.10)}&19.65\textcolor{blue}{(10.42)}& \cellcolor{gray!20}3.85 \\
$\ell_1$-sparse &96.07\textcolor{blue}{(1.69)}&99.89\textcolor{blue}{(0.34)}&96.65\textcolor{blue}{(1.04)}&7.80\textcolor{blue}{(1.43)}& \cellcolor{gray!20}1.13 \\
SalUn &93.78\textcolor{blue}{(0.60)}&96.61\textcolor{blue}{(2.94)}&94.19\textcolor{blue}{(1.42)}&31.00\textcolor{blue}{(21.77)}& \cellcolor{gray!20}6.68 \\
LetheViT &96.02\textcolor{blue}{(1.64)}&99.65\textcolor{blue}{(0.10)}&96.84\textcolor{blue}{(1.23)}&7.73\textcolor{blue}{(1.50)}& \cellcolor{gray!20}1.12 \\

\bottomrule
\end{tabular}
\caption{Performance of various MU methods for DeiT-S on SVHN. The unlearning scenarios include 10\%, 30\%, and 50\% forgetting rates. \textbf{Bold} indicates the best AG.}
\label{tab10}
\end{table*}

\end{document}